\documentclass[sigconf,natbib=true]{acmart}

\usepackage{soul}
\usepackage{graphicx}
\usepackage{subfig}
\usepackage{multirow}
\graphicspath{{figures/}}
\usepackage{enumitem}
\newlist{romanenum}{enumerate}{1}
\setlist[romanenum]{label=(\roman*)} 
\newlist{rqenum}{enumerate}{1}
\setlist[rqenum]{label=[\textbf{RQ\arabic*}]}

\definecolor{lightblue}{rgb}{.50,.95,1}
\definecolor{tri}{rgb}{.25,.88,.82}
\definecolor{lilac}{rgb}{0.85,0.64,0.85}

\newcommand{\ds}{CT--CWT--21}
\newcommand{\abt}{AraBERTv1}
\newcommand{\bt}{BERT}
\newcommand{\bturk}{BERTurk}
\newcommand{\slbt}{SlavicBERT}
\newcommand{\spbt}{BETO}
\newcommand{\mbt}{mBERT}
\newcommand{\monbt}{monoBERT}
\newcommand{\zs}{ZS}
\newcommand{\tr}{ZS-Tr}
\newcommand{\trzs}{ZS-TrSrc}
\newcommand{\tezs}{ZS-TrTrg}
\newcommand{\advzs}{ZS-Adv}
\newcommand{\fs}{FS}
\newcommand{\cw}{check-worthiness}
\newcommand{\bemph}[1]{\emph{\textbf{#1}}}
\usepackage{color}
\usepackage{colortbl}

\setcopyright{acmcopyright}
\copyrightyear{2022}
\acmYear{2022}
\acmDOI{XXXXXXX.XXXXXXX}
\acmPrice{15.00}
\acmISBN{978-1-4503-XXXX-X/18/06}

\acmConference[SIGIR '22]{the 45th International ACM SIGIR Conference on Research and Development in Information Retrieval}{July 11--15, 2022}{Madrid, Spain}
  
\acmBooktitle{Proceedings of the 45th International ACM SIGIR Conference on Research and Development in Information Retrieval (SIGIR '22), July 11--15, 2022, Madrid, Spain}

\begin{document}


\title{Cross-lingual Transfer Learning for Check-worthy Claim Identification over Twitter}

\author{Maram Hasanain and Tamer Elsayed}
\affiliation{%
  \institution{Qatar University}
  \city{Doha}
  \country{Qatar}}
\email{{mh081131,telsayed}@qu.edu.qa}

\renewcommand{\shortauthors}{M. Hasanain and T. Elsayed}

\begin{abstract}
Misinformation spread over social media has become an undeniable infodemic. However, not all spreading claims are made equal. If propagated, some claims can be destructive, not only on the individual level, but to organizations and even countries. Detecting claims that should be prioritized for fact-checking is considered the first step to fight against spread of fake news. With training data limited to a handful of languages, developing supervised models to tackle the problem over lower-resource languages is currently infeasible. Therefore, our work aims to investigate whether we can use existing datasets to train models for predicting worthiness of verification of claims in tweets in other languages. We present a systematic comparative study of six approaches for cross-lingual check-worthiness estimation across pairs of five diverse languages with the help of Multilingual BERT (\mbt) model. We run our experiments using a state-of-the-art multilingual Twitter dataset. Our results show that for some language pairs, zero-shot cross-lingual transfer is possible and can perform as good as monolingual models that are trained on the target language. We also show that in some languages, this approach outperforms (or at least is comparable to) state-of-the-art models.   
\end{abstract}



\maketitle

\section{Introduction}
\label{intro}
\begin{figure}[t]
  \centering
  \subfloat[a][Check-worthy claim: ``(translation) Jordan denies entry of Chinese citizens due to the Corona virus."]{\includegraphics[width=0.7\columnwidth]{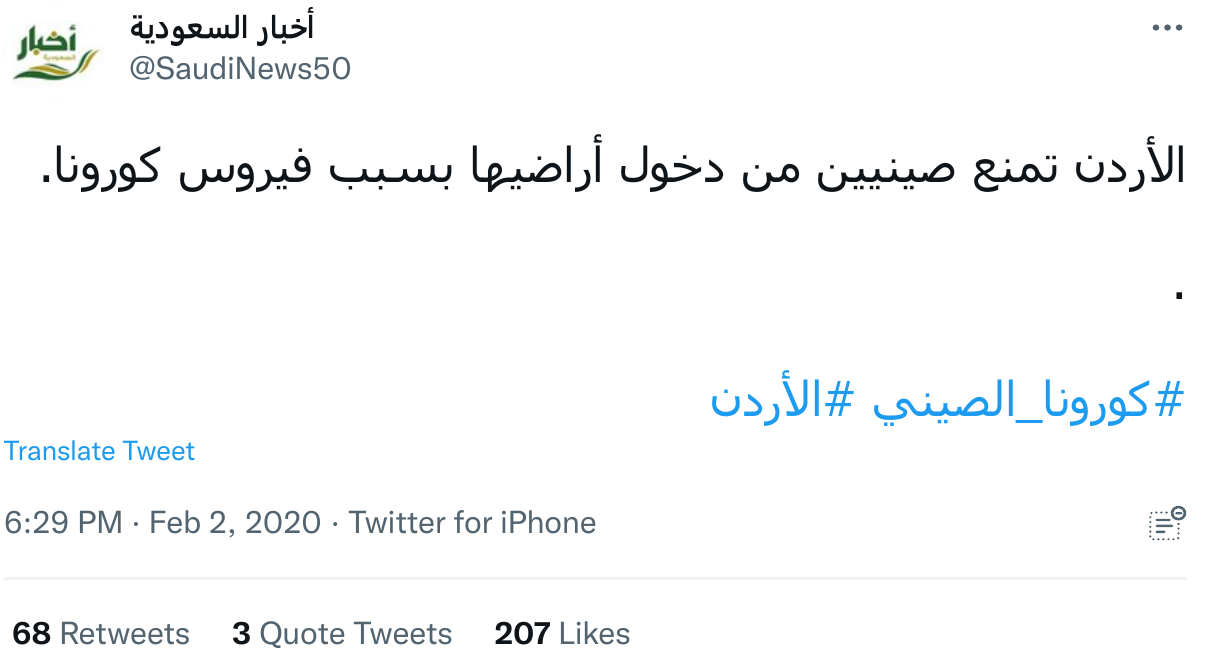} \label{ex1:ar:cw}} \\
  \subfloat[b][Non-check-worthy claim: ``(translation) Ghassan Salamé, head of the United Nations support mission in Libya holds a press conference. "]{\includegraphics[width=0.7\columnwidth]{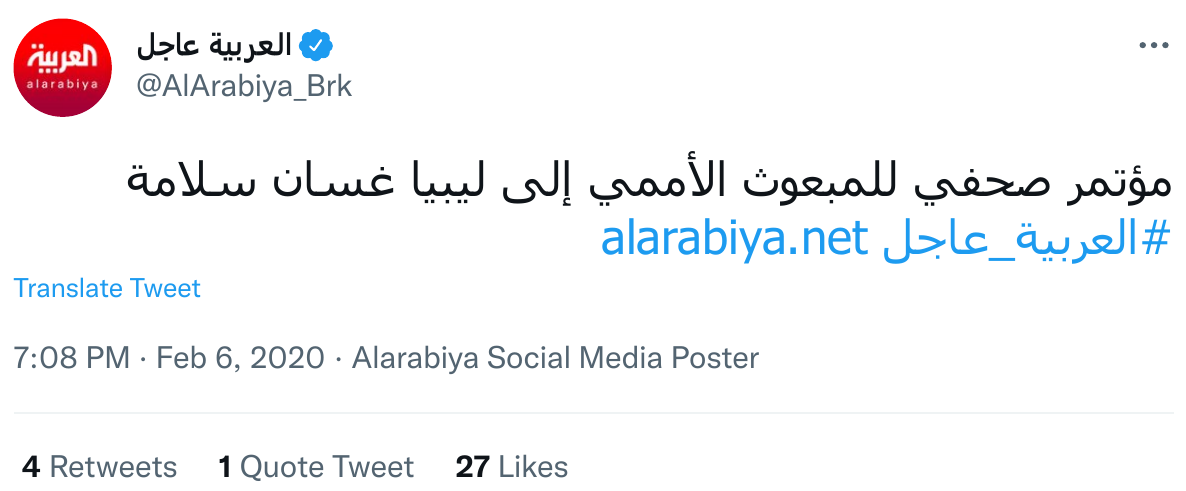} \label{ex1:ar:ncw}}
  \caption{An example comparing check-worthy and none check-worthy Arabic claims (with translation)}     \label{ex1:ar}
\end{figure}

One of the major issues faced by typical users of social media is misinformation infecting their timelines. 
Manual and automated efforts to detect and verify claims are indispensable to protect and inform users, especially in critical times like the current COVID-19 pandemic~\cite{brennen2020,nakov2021automated}. While scanning a timeline, a user or a fact-checker is faced by many posts that are potentially false. Verifying all these claims can become cumbersome. Thus, the first step in the process of fact checking is identifying which posts contain claims that are worth verifying~\cite{nakov2021automated}. Not all claims are as important; some can have catastrophic impact on a large population, such as the popular claims discouraging COVID-19 vaccination~\cite{sear2020quantifying}. Other claims might not cause any lasting impact or invoking any action. Figure~\ref{ex1:ar} shows examples of tweets containing claims borrowed from Task 1 of the CheckThat! 2021 evaluation lab~\cite{clef-checkthat:2021:task1}. Tweet in Figure~\ref{ex1:ar:cw} was labelled as containing a check-worthy claim since the news might cause an international political crisis if propagated, making verification of the claim essential before its spread. 

A claim check-worthiness is usually defined by its importance to the public and fact-checkers~\cite{Hassan2017,arslan2020benchmark,clef-checkthat:2021:task1}. However, for this work, we adopt a more concrete definition from the check-worthiness estimation task at the CheckThat! lab at CLEF2021~\cite{clef-checkthat:2021:LNCS,clef-checkthat:2021:task1}. In the task, a check-worthy sentence is one that: 1) contains a factual claim, 2) is of interest to the public, 3) can potentially cause emotional or physical harm to a person or an organization, 4) a journalist might be interested in covering 
, and 5) a fact-checker should verify.

Social media enables and contains highly multilingual streams with many users following content in multiple languages. Moreover, specific groups of users such as journalists or news agencies usually track multilingual content for news reporting and even fact-checking~\cite{fullfact:coof}. The same claim might propagate in multiple languages (e.g., after translation) which might result in repeated verification efforts across languages~\cite{fullfact:coof,nakov2021automated}. This poses a need for multilingual systems for check-worthy claim detection. Most existing systems are monolingual, such as those participating in the CheckThat! challenge~\cite{clef-checkthat:2021:task1}. Recently, few exceptions emerged, usually considering multilingual transformer models to handle multilingual input (e.g.,~\cite{clef-checkthat:2021:task1:Schlicht2021,uyangodage2021}). Such preliminary studies focused on multilingual support, i.e., training a system using multilingual data including the targeted language and testing it on a set of sentences in a target language. This setup assumes that some training data is available for the target language. However, for low resource languages and for an emerging problem, such as the check-worthiness detection, this might not be feasible. 

In this work, we ask the following question: \bemph{can we build an effective supervised model for check-worthiness detection without the need for training data in the target language}? We address this question by testing six setups to perform cross- and multi-lingual check-worthiness detection over tweets. Our work mainly focuses on a well-known setup in related problems, called \textit{zero-shot} transfer learning, where no labeled examples in the target language are used during model training or fine-tuning. We start from the highly effective classification architecture based on multilingual \bt{} (\mbt)~\cite{devlin2019}. Architectures based on \mbt{} demonstrated effectiveness in cross-lingual transfer learning in several text classification tasks~\cite{wu2019beto}. Up to our knowledge, this is the first study of this kind and scale for the problem of check-worthiness detection. We mainly address the following research questions:
\begin{rqenum}[leftmargin=1.2cm]
\itemsep0em
\item Given labeled data in a source language, how effective is zero-shot cross-lingual check-worthiness prediction on a different target language?
\item Does translation between source and target languages improve the performance?
\item How much improvement can we achieve by adding few labeled examples in the target language to labeled examples in source language (i.e., few-shot transfer learning)?
\item Will adversarial training with unlabeled examples in target language improve over the zero-shot cross-lingual setup? 
\item Can we improve the performance if we transfer from multiple source languages to a single target language?
\item How effective is cross-lingual transfer compared to the state of the art models?
\end{rqenum}

Our contribution in this work is four-fold: 
\begin{itemize}
    \item We extensively explore and benchmark diverse methods to train cross-lingual check-worthiness prediction models including zero-shot, few-shot, and translation-based approaches. Existing studies for the task have not provided such a large-scale comparative study with different variants.
    \item We demonstrate that for some language pairs, cross-lingual transfer learning (e.g., from Arabic to Turkish) is at least comparable or even significantly better than monolingual models exclusively trained on target language. While for other languages (Bulgarian and Spanish), cross-lingual transfer is not effective regardless of the setup or source language.
    \item Our results show that for some target languages, cross-lingual transfer models are as good as state-of-the art models developed for check-worthiness estimation on the same dataset.  
    \item This study is the first to experiment with adversarial training for cross-lingual check-worthiness prediction. 
\end{itemize}



The remainder of this paper is presented as follows. We summarize existing studies in Section~\ref{related}. Section~\ref{cross_ling_cw} presents the approach and design of experiments we follow in this study. We present and discuss the experimental setup and results in Sections~\ref{impl} and \ref{results}. Section~\ref{conc} summarizes the work and presents concluding remarks.
\section{Related Work}
\label{related}
The problems of claim detection and verification has attracted enormous attention in the past few years. Due to the volume of proposed systems, several literature surveys already exist (e.g.,~\cite{guo2021survey,nakov2021automated}). Thus, we only focus here on two aspects: check-worthiness detection over tweets in general and  multilingual approaches to the problem.

ClaimBuster is one of the pioneering approaches to check-worthiness detection~\cite{Hassan2017}. The system computes features for each input sentence such as its sentiment score, and part-of-speech tags and trains a supervised model with typical classifiers (e.g., SVM). More recent systems usually use neural models and specifically, classification architectures based on transformer models (e.g.,~\cite{Martinez-Rico2021,Sepulveda2021,clef-checkthat:2021:task1:accenture,beltran2021claimhunter,alhindi2021fact,wright2020claim}). A more recent version of ClaimBuster~\cite{meng2020gradient} combines \bt{}~\cite{devlin2019} and gradient-based adversarial training to build a more effective model. In this system, perturbations are added to the embeddings generated by \bt{} for an input sentence, and the final model is fine-tuned minimizing both classification and adversarial losses. Both ClaimBuster models were tested on political debates or tweets but limited to English, while we target multilingual streams.  

In a very recent study~\cite{Alam_2021}, a dataset of English and Arabic tweets about COVID-19 was annotated on several aspects including check-worthiness. The authors fine-tune several transformer models for the task, but train a model for each of Arabic and English independently. In a further study~\cite{alam2021-fighting-covid}, the dataset was augmented with Bulgarian and Dutch tweets, and initial experiments on multilingual classification were conducted. In the proposed system, \mbt{} model~\cite{devlin2019} is fine-tuned using all of the four languages and then tested on each. Uyangodage et al.~\cite{uyangodage2021} follow the same approach but considering two datasets: NLP4IF~\cite{shaar-etal-2021-findings} and the CheckThat! 2021 Task 1 dataset~\cite{clef-checkthat:2021:task1}. Differently and more comprehensively, we examine multiple alternatives for cross-lingual transfer where minimal or no training data in the target language is required. Moreover, we do not limit our work to tweets on one topic (i.e., COVID-19).

Among the most prominent efforts to approach the problem of check-worthiness detection are those part of the CLEF CheckThat! lab for the past four years~\cite{clef-checkthat:2021:LNCS}. In the initial two editions of the lab, the problem targeted claims within political debates~\cite{clef-checkthat-T1:2018,clef-checkthat-T1:2019}. In the next editions, the lab focused on the social media domain and specifically, check-worthiness estimation for tweets~\cite{clef-checkthat-ar:2020,clef-checkthat-en:2020,clef-checkthat:2021:task1}. The problem was defined as follows: given a stream of tweets on a topic, the participating systems were asked to rank the tweets by check-worthiness for the topic. Our work focuses on a more general definition of the problem, modeling it as a classification task without a limitation to any topic. That is to say, we aim to develop a system to detect check-worthy claims in a general stream of tweets. This definition is inline with some of the existing studies~\cite{Hassan2017,arslan2020benchmark,meng2020gradient}.

The last lab edition (CheckThat! 2021) offered a first-of-its-kind multilingual dataset (\ds) for the problem. The dataset contained labelled tweets in five languages: Arabic, Bulgarian, English, Spanish and Turkish. This is the evaluation dataset we use in this work (further details in Section~\ref{datasets}). Only few systems participating in the lab attempted to benefit from the unique nature of this dataset. Schlicht et al. (team UPV)~\cite{clef-checkthat:2021:task1:Schlicht2021} proposed a transformer-based model jointly trained for two tasks: check-worthiness detection and language identification. The team fine-tunes a multilingual transformer model called sentence-BERT~\cite{reimers2020making} optimizing for both classification tasks. The language identification task aims at mitigating bias to any of the training languages. Again, in their study, authors train the model over all five languages in \ds, however, we focus on cross-lingual transfer. The work of Zengin et al.~\cite{tobbetu2021} is the closest to ours. The authors attempted a cross-lingual approach where \mbt{} is fine-tuned on each pair of the five languages in \ds, then tested per language. Differently, we examine a wider set of variants for cross-lingual check-worthiness estimation and show how they compare to several existing baselines. In a more recent work by the same authors~\cite{Kartal2022}, \mbt{} is tested in cross-lingual transfer for three languages only (Arabic, English and Turkish). We also observe a potential source of issues in their evaluation setup since the datasets came from different domains (political debates and tweets) and follow different annotation strategies across the languages, while in our setup we maintain consistency as much as possible using tweets only.
 
\section{Approach}
\label{cross_ling_cw}
Our study tackles the problem of check-worthiness detection over multilingual streams. We formally define the problem as follows: \emph{Given a stream of tweets in any language, detect tweets containing check-worthy claims.} This section describes the main architecture we use throughout our experiments. We then present the design of experiments answering our research questions. 
\subsection{System Architecture}
\label{arch}
Our work is motivated by the strong line of research showing the effectiveness of transformer models, such as \bt{}, for text classification. In the area of fact-checking, architectures based on transformer models are among the best performing for different tasks including \cw{} prediction~\cite{clef-checkthat:2021:task1,accenture2021,tobbetu2021,Martinez-Rico2021}, claim verification~\cite{zeng:survey:2021} and evidence retrieval~\cite{samarinas-etal-2021-improving}.

\begin{figure}[t]
    \centering
    \includegraphics[width=0.4\columnwidth]{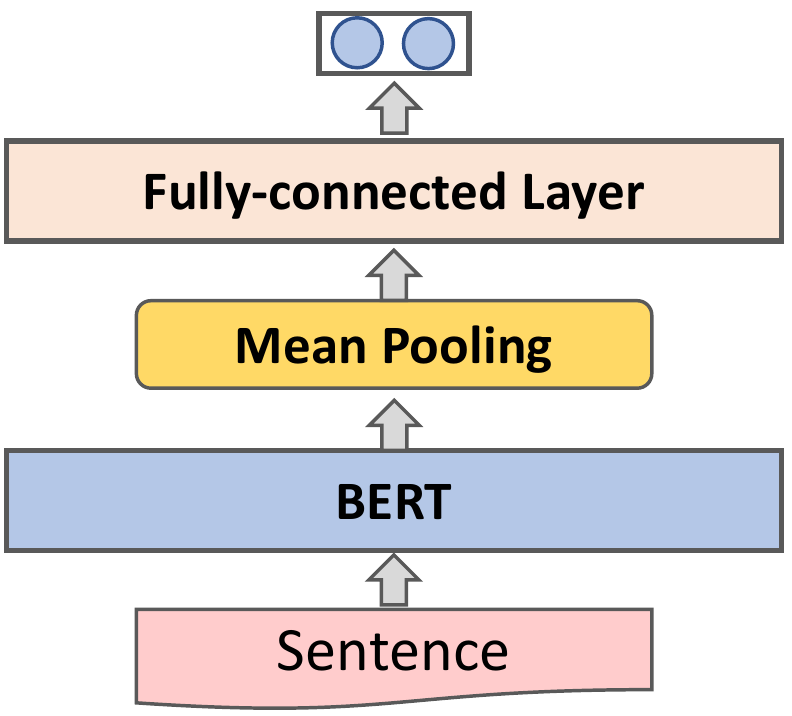}
    \caption{Classification architecture. \bt{} layer represents all \bt-based transformer models used in this work.}
    \label{fig:arch}
\end{figure}

For all our experiments, we start from the same \bt-based classification architecture depicted in Figure~\ref{fig:arch}. This architecture is constructed based on previous literature using \bt{} for text classification. Specifically, following \bt{} layers, we add a feed-forward network with one hidden linear layer (of 256 nodes) with ReLu activation. Softmax activation function is finally applied to the output layer, resulting in two predicted probabilities (one for each of the two classes). As an input, we pass a single sequence which is the sentence \bemph{S} that we would like to predict its \cw. The input to our model is formatted as follows: [[CLS], \bemph{S}, [SEP]]. Typically, after training the full architecture (including fine-tuning of the pre-trained model), the hidden state \textbf{h} produced by the transformer model for the [CLS] token is used as representation of the input to the remainder of the classification architecture. However, during our preliminary experiments on development subsets, we found that using mean pooling over all tokens yields better classification results, thus we adopt this pooled representation. At inference time, the probability of the positive class determines the predicted label for the input sentence with a 0.5 threshold. 
We train the model minimizing cross entropy loss.

\subsection{Cross-lingual Check-worthiness Transfer}
The main aim of our work is to investigate whether \cw{} learning can be transferred across languages, and then identify potentially effective systems with none or minimal labeled data originally written in a target language. To that end, we study different strategies for transfer learning from source language \emph{A} to target language \emph{B}, starting from the zero-shot transfer learning setup, going through methods that employ minimal labelled (or unlabelled) data from the target language and domain, and finally, approaches that enrich zero-shot transfer learning with translation. We next describe each of the approaches investigated in this work. 

\subsubsection{Zero-shot Cross-lingual Transfer Learning (\textbf{\zs})}
Given the strong ability of pre-trained models, such as, \mbt{} in zero-shot cross-lingual transfer over multiple NLP tasks~\cite{wu2019beto}, we fine-tune the system described in Section~\ref{arch} on a source language and apply it directly at inference time to a test set in the target language. This approach represents the basic \zs{} model we test in this work; it is a commonly-adopted approach in related cross-lingual transfer learning studies~\cite{zhao2021closer}. 

\subsubsection{Zero-shot with Translation (\textbf{\tr})}
In the second research question, we aim to find an improved setup over \zs{} by translation.\footnote{We used Google Translation API at \url{https://cloud.google.com/translate.}} Instead of depending on transfer ability of \mbt, we unify the language of both train and test sets using two strategies:
\begin{itemize}
    \item \textbf{\trzs}: 
    In this setup, we translate the \emph{training} set of \bemph{source} language \emph{A} to the target language \emph{B}, then fine-tune our model on the translated data. The model is then directly applied to the test set of the target language \emph{B}.
    \item \textbf{\tezs}: 
    This setup shows the second possible translation approach. We first fine-tune our model using the \textit{original} training set of the source language A. We then translate the \emph{test} set of the \bemph{target} language \emph{B} into language \emph{A}, and apply our model on it.
\end{itemize}


\subsubsection{Transfer Learning with Few Shots (\textbf{\fs})}
In this setup, we experiment with transfer learning \textit{extended} with the addition of few labelled training examples from the training set of the target language. This is different from the translation-based approaches, since the labeled examples where originally written in the target language, rather than being translated. Few-shot cross-lingual transfer with  \emph{two-stage} fine-tuning has gained importance recently, since it generally improves performance with small annotation cost for target language examples (e.g., ~\cite{hedderich2020}). In this setup, we fine-tune our model in two stages; first it is fine-tuned over the source language \emph{A}, then further fine-tuned using few examples from the target language \emph{B}. We use random sampling with balanced classes to select few shots for the second stage (details in Section~\ref{fs:results}).

\subsubsection{\zs{} with Adversarial Training (\textbf{\advzs})}
\label{sec:adv}
Adversarial learning is one the approaches that was shown to be successful in zero-shot cross-lingual transfer learning for text classification~\cite{keung2019,dong2020leveraging}. Existing methods generally used adversarial networks and training techniques to extract features that are invariant to the change of language when performing cross-lingual transfer~\cite{chen2019multi}. We test the effect of using adversarial training to improve \zs{} for our problem. To that end, we test an existing model~\cite{keung2019} proposed for other text classification tasks, and test it only using English as the source language. This model extends typical \mbt{} classification architecture with the addition of an adversarial classification task trained in parallel to the actual  classification task (\cw{} in our case). This additional task includes two components: a discriminator that uses \mbt{} embeddings representing an input sentence to predict whether it is in the source language, and a generator that tries to generate embeddings that are difficult for the discriminator to detect. Eventually, this adversarial task aims to train \mbt{} to generate embeddings less fitted to the source training language while being fine-tuned for the text classification task. 

To train such architecture, we need two components: training set in the source language and unlabeled sentences from the target language. In our experiments (Section~\ref{results:adv}), we use unlabeled examples sampled from the target training set which results in more consistency between training and test sets in terms of domain and topics. However, we note that this is not an optimal scenario as this might not be easy to achieve in the case of some target languages where a training set in that language is not available. We leave investigating the use of other approaches to acquire sentences in the target language to future work.
\section{Experimental Setup}
\label{impl}
This section presents the experimental evaluation setup, designed to answer our research questions, including the datasets used in experiments, evaluation approach, and implementation details for the classification architecture.

\subsection{Datasets}
\label{datasets}
Before running this study, we needed to identify a dataset that allows us to examine check-worthiness estimation in a multilingual setup. Since this problem is very recent, we only found two datasets that satisfy this condition. The first covers 4 languages (Arabic, Bulgarian, Dutch, and English) and limited only to tweets about COVID-19~\cite{Alam_2021}. We opt to use the second one (denoted as \ds) that was part of the CheckThat! lab of CLEF 2021 conference~\cite{clef-checkthat:2021:LNCS}. The dataset was created to evaluate systems for the task of check-worthiness estimation over tweets in five languages, namely Arabic, Bulgarian, English, Spanish, and Turkish. In addition to having five languages as opposed to four in~\cite{Alam_2021}, \ds{} is not limited to COVID-19 only. Generally, tweets in \ds{} cover two general topics, politics and COVID-19.  In \ds{}, the per language subsets were created independently following a similar definition of check-worthiness. A tweet in each subset has two potential labels: check-worthy claim or non-check-worthy sentence. We consider the former label to be the positive label in this work. 

Before proceeding with \ds, we first combine the training and development sets per language to acquire a larger training set. We observed a great difference in the training set size across languages, ranging from 962 to 4.1k tweets. More importantly, the class distribution varies significantly with the percentage of positive labels falling between 8\% and 38\% across languages. Although such class distribution prior might be observed in real-world cases, this can shift the focus of this work from understanding check-worthiness estimation differences across languages to how to best handle this imbalance. Moreover, this imbalance can mask or exaggerate system performance across languages. Such observations were made in previous research concerning systems for cross-lingual transfer~\cite{SCHWENK18.658}.

We alleviate the problem of varied dataset sizes across languages by down sampling the training subset per language using a stratified random sampling approach. This ensures that we have the same dataset size and class distribution across languages. We chose the sample size per class based on the minimum number of labels per class across languages. Eventually, for each language except English, we end-up with 300 positive and 1,400 negative examples. As for the English dataset, number of negative examples available was smaller than 1,400. Thus, we have 300 positive and only 612 negative \emph{English} examples. As for the test subsets, we keep them as released in the lab to allow for comparison with existing systems tested on the same subsets. Moreover, our aim in this work is not to compare the system performance on the test subsets across languages, and thus, variation in distributions of labels across test subsets does not affect the conclusions we make in this study. Table~\ref{tab:datasets} summarizes datasets per language.

\begin{table}[t]
  \caption{Datasets used in experiments.}
  \label{tab:datasets}
  \small
  \begin{tabular}{llcccc}
    \toprule
               \multicolumn{ 2}{c}{} & \multicolumn{ 2}{c}{{\bf Train}} & \multicolumn{ 2}{c}{{\bf Test}} \\
\hline
{\bf Dataset} & {\bf Topics} & {\bf Total} & {\bf \# CW} & {\bf Total} & {\bf \# CW} \\
\hline
Arabic & Politics \& COVID-19 &  1700 &   300 &   600 &   242 \\
Bulgarian & COVID-19 &  1700 &   300 &   357 &    76 \\
English & COVID-19 &   912 &   300 &   350 &    19 \\
Spanish &  Politics &  1700 &   300 &  1248 &   120 \\
Turkish &  Politics \& COVID-19 &  1700 &   300 &  1013 &   183 \\
  \bottomrule
\end{tabular}
\end{table}

\subsection{Implementation Details}
Due to the extent of the experiments we run and the limited dataset sizes, we unify the model parameters across experiments without hyperparameter tuning (unless otherwise stated). We set the parameters following optimal values identified in the original \bt{} paper (Appendix A.3 in ~\cite{devlin2019}) and a recent paper that examined \mbt{} performance for multilingual text classification~\cite{pires2019}. We set the training batch size to 32, learning rate to 3e-5, and a maximum sequence length of 128. We fine-tuned the model for three epochs (in line with related work on the same dataset~\cite{uyangodage2021}) and repeated model training five times with different random seeds to account for any randomness in model initialization and training. In this work, we report the average performance over those five re-runs.

We evaluate the models using the $F_1$ score of predicting the positive class. We chose this measure since our aim is to understand the model effectiveness in identifying check-worthy claims. The statistical significance of difference between systems is tested with two-sided paired t-test over the five re-runs with $\alpha<0.05$.

For all experiments, we use pre-trained \bt{} models from the HuggingFace (HF) library (Table~\ref{tab:models}).\footnote{\url{https://huggingface.co/models}} Base version of the models with 12 layers was used. These models are among the most effective and commonly-used for the target languages. For all languages but Bulgarian, we found a model solely pre-trained on the target language. For Bulgarian, we only found a model pre-trained on four Slavic languages including Bulgarian.

\begin{table}
  \caption{Pre-trained models used in experiments.}
  \label{tab:models}
    \small
  \begin{tabular}{lll}
    \toprule
{\bf Model} & {\bf Language} & {\bf HF Model Name} \\
\hline
\mbt{}~\cite{devlin2019} & Multilingual & bert-base-multilingual-cased \\
\abt{}~\cite{antoun2020arabert} & Arabic & aubmindlab/bert-base-arabert \\
\bt{}~\cite{devlin2019} &    English & bert-base-uncased \\
\slbt{}~\cite{arkhipov-etal-2019-tuning} &    Slavic & DeepPavlov/bert-base-bg-cs-pl-ru-cased \\
\spbt{}~\cite{CaneteCFP2020} & Spanish & dccuchile/bert-base-spanish-wwm-cased\\
\bturk{}~\cite{stefan_schweter_2020_3770924} & Turkish & dbmdz/bert-base-turkish-cased\\
\bottomrule
\end{tabular}
\end{table}

\section{Results and Discussion}
\label{results}
In this section, we present and discuss the results of the experiments we designed to answer each of the research questions. 

As a \emph{baseline} for all of our experiments (unless otherwise stated), we report the performance of \mbt{} when fine-tuned on the training set of the target language.

\subsection{Zero-shot Cross-lingual Transfer Learning}
\label{result:zero-shot}
We start by understanding the model effectiveness in zero-shot cross-lingual transfer learning (\textbf{RQ1}). For this purpose, we train an independent model for each of the five languages, then report the models' performance on the test set of each target language. 
Table~\ref{tab:rq1_zero} shows the results, and the baseline per language on the diagonal. 

\begin{table}
  \caption{\zs{} results. \textbf{Bold} and \underline{underlined} values represent best and second best per test set, respectively. $^{*}$ indicates significant difference from baseline on same test set.}
  \label{tab:rq1_zero}
    \small
  \begin{tabular}{l|ccccc}
    \toprule
    & \multicolumn{5}{c}{\bf target}\\
    \cline{2-6}
\multirow{2}{*}[1em]{\bf source} & \multicolumn{1}{c}{\bf ar} & \multicolumn{1}{c}{\bf bg}& \multicolumn{1}{c}{\bf en} & \multicolumn{1}{c}{\bf es} & \multicolumn{1}{c}{\bf tr}\\
\hline
  {\bf ar} & {\cellcolor [gray]{0.8}}{\textbf{49.5}} & 35.2$^{*}$ & 12.1$^{*}$ & 26.6$^{*}$ & \textbf{50.3}$^{*}$\\
  {\bf bg} & 04.6$^{*}$ & {\cellcolor [gray]{0.8}}\textbf{58.2} & \underline{16.6}$^{*}$ & 07.4$^{*}$ & 27.4 \\
  {\bf en} & \underline{47.7} & \underline{41.0}$^{*}$ &  {\cellcolor [gray]{0.8}}13.3 & \underline{27.6}$^{*}$ & \underline{46.1}$^{*}$ \\
  {\bf es} & 00.3$^{*}$ & 06.7$^{*}$ & 11.1 & {\cellcolor [gray]{0.8}}\textbf{54.0} & 24.8 \\
  {\bf tr} & 12.3$^{*}$ & 20.5$^{*}$ & \textbf{16.8} & 15.6$^{*}$ & {\cellcolor [gray]{0.8}}28.4 \\
\bottomrule
\end{tabular}
\end{table}

This experiment shows promising results that answer \textbf{RQ1}. We notice that for Arabic, English, and Turkish, the best \zs{} performance is at least as good as the baseline where \mbt{} was fine-tuned on the target language. We even observe that for English and Turkish, we identified a source language that results in a significant improvement over the baseline. One of the most notable observations in Table~\ref{tab:rq1_zero} is the great performance improvement due to \zs{} from Arabic to Turkish. We anticipate this is the case for at least the following reason. All language subsets, except Spanish (\textbf{es}), had similar annotation strategy. However, we observe that \textbf{en} and \textbf{bg} datasets only focused on tweets about COVID-19, while \textbf{ar} and \textbf{tr} covered other additional topics, making generalization from \textbf{ar} to \textbf{tr} more probable. One might stop here and notice that the transfer in the opposite direction from \textbf{tr} to \textbf{ar} is ineffective compared to the monolingual baseline. We believe this is due to the pre-training of \mbt{}, as Turkish is a lower resource language that was underrepresented compared to Arabic. Such issue has been shown to negatively affect language representation learnt by \mbt{} and thus affect performance of the model after fine-tuning~\cite{wu2020all}. A similar observation holds for another lower resource language, namely Bulgarian, as shown in the table. \emph{Overall, this experiment showed that for the check-worthiness prediction task, zero-shot transfer is as effective as fine-tuning \mbt{} over full training sets of three target languages: Arabic, English and Turkish.}   

\subsection{Effect of Translation on \zs}
\label{result:trans}
In \textbf{RQ2}, we aim to find an improved setup over \zs{} by translation. In this experiment, we show check-worthiness estimation performance when translating from each of the two directions: translate from the source language or from the target language.
\subsubsection{\zs{} with Source Translation (\trzs)}
\label{result:transtrain}
Table~\ref{tab:rq2_tr1} shows results with translation applied to the training set of the source language, with the baseline per language on the diagonal. We find that, on average, translation of the source language resulted in improved performance over original \zs{} in 8 out of 20 pairs, with an average improvement of 5.7 points in $F_1$. As for the cases when performance degradation is observed, the average of the absolute differences between baseline and translation setups is 3.1 points in $F_1$. Overall, that indicates slight improvement with translation. Translation did not seem to help Turkish specifically as the performance degraded with all source languages. For the remaining target languages, pairs of languages had varied performance. This can be the result of the effectiveness of translation system used.

\begin{table}
  \caption{Effect of translating the source language on transfer learning. Values in () represent the percentage of difference between the performance in this setup and corresponding \zs{} cell from Table~\ref{tab:rq1_zero}. $^{*}$ indicates significant difference from baseline on same target language.}
  \label{tab:rq2_tr1}
    \small
 \begin{tabular}{l|lllll}
    \toprule
    & \multicolumn{5}{c}{\bf target}\\
    \cline{2-6}
\multirow{2}{*}[1em]{\bf source} & \multicolumn{1}{c}{\bf ar} & \multicolumn{1}{c}{\bf bg}& \multicolumn{1}{c}{\bf en} & \multicolumn{1}{c}{\bf es} & \multicolumn{1}{c}{\bf tr}\\
\hline
  {\bf ar} & {\cellcolor [gray]{0.8}}\textbf{49.5} & 31.2$^{*}$ (-11)  & 10.3$^{*}$ \space(-15)  & 25.9$^{*}$ (-3)  & \textbf{47.0} (-7)  \\
  {\bf bg} & 09.7$^{*}$ (111)  &  {\cellcolor [gray]{0.8}}\textbf{58.2} & \textbf{17.9$^{*}$} (8)  & 19.2$^{*}$ (159)  & 21.8 (-20) \\
  {\bf en} & \underline{45.7} \space\space(-4)  & \underline{36.3$^{*}$} (-11) &  {\cellcolor [gray]{0.8}}13.3 & \underline{30.0$^{*}$} (9) & \underline{38.0} (-18)\\
  {\bf es} & 00.0$^{*}$ (-100)  & 17.3$^{*}$ (158)  & 16.0 \space\space(44)  & {\cellcolor [gray]{0.8}}\textbf{54.0} & 24.6 (-1) \\
  {\bf tr} & 18.9$^{*}$ (54)  & 14.0$^{*}$ (-32)  & \underline{16.3} \space\space(-3) & 18.4$^{*}$ (18)  & {\cellcolor [gray]{0.8}}28.4 \\
\bottomrule
\end{tabular}
\end{table}

\subsubsection{\zs{} with Target Translation (\tezs)}
\label{result:transtest}
We continue to answer \textbf{RQ2} by translation of the test set of the target language to match the source language. Table~\ref{tab:rq2_tr2} shows results under this setup. We find that, on average, translation of the target language resulted in improved performance over original \zs{} in 10 out of 20 pairs, with an average performance improvement of 3.8 points in $F_1$.  As for the cases when performance degradation is observed, the average of the absolute differences between baseline and translation setups is 1.6 points in $F_1$. Overall, that indicates slight improvement with translation.

\begin{table}
  \caption{Effect of translating the target language on transfer learning. Values in () represent the percentage of difference between the results for this setup and corresponding \zs{} cell from Table~\ref{tab:rq1_zero}.$^{*}$ indicates significant difference from baseline on same target language.}
  \label{tab:rq2_tr2}
    \small
 \begin{tabular}{l|lllll}
    \toprule
    & \multicolumn{5}{c}{\bf target}\\
    \cline{2-6}
\multirow{2}{*}[1em]{\bf source} & \multicolumn{1}{c}{\bf ar} & \multicolumn{1}{c}{\bf bg}& \multicolumn{1}{c}{\bf en} & \multicolumn{1}{c}{\bf es} & \multicolumn{1}{c}{\bf tr}\\
\hline
  {\bf ar} &{\cellcolor [gray]{0.8}}\underline{49.5} & 35.0$^{*}$ (-1) & 12.0 \space(-1)  & 25.9$^{*}$ (-3) & \textbf{51.9}$^{*}$ (3) \\

  {\bf bg} & 05.9$^{*}$ (28) & {\cellcolor [gray]{0.8}}{\bf 58.2} & {\bf 14.1} (-15) & 11.0$^{*}$ (49) & 34.4 \space\space(26)  \\
  
  {\bf en} & {\bf 52.0} \space(9) & \underline{40.9$^{*}$} (0) & {\cellcolor [gray]{0.8}}\underline{13.3} & \underline{27.9$^{*}$} (1)  & \underline{51.6$^{*}$} (12)\\
  
  {\bf es} & 01.5$^{*}$ (400)  & 06.5$^{*}$ (-3)  & 08.5 \space(-23)  &{\cellcolor [gray]{0.8}}{\bf 54.0} & 28.5 \space\space(15)  \\
  
  {\bf tr} & 21.8$^{*}$ (77)  & 19.0$^{*}$ (-7)  & 10.3 \space(-39)  & 14.3$^{*}$ (-8) &{\cellcolor [gray]{0.8}}28.4 \\
\bottomrule
\end{tabular}
\end{table}

To further understand why translation might help in some cases over \zs, we dive into the actual predictions made in these two setups, taking transfer from \textbf{en} to \textbf{ar} as an example. For this experiment, we select the run (out of the five re-runs) that achieved the best performance improvement compared to the corresponding \zs{} run. We found that 60 tweets were correctly predicted by \tezs{} but miss-classified by \zs. Two thirds of these tweets discuss a single topic ``2021 United States Capitol attack''. This explains why translation from \textbf{ar} test set to \textbf{en} helped improve prediction over these tweets that are directly related to U.S. politics. Vocabulary of this topic will more probably appear in English Wikipedia, versus the Arabic version. Wikipedia was used to pre-train the \mbt{} model with English being the most represented language. Thus, the model is more fitted to the language and topics used in English Wikipedia, making it easier to classify these Arabic tweets when translated to English.
\emph{The initial analysis of tweets in this case raised an important questions about the check-worthiness prediction task definition itself. What's the effect of claim topic on model transfer? and should the topic be considered as part of the task definition to acquire a better context of the claim?}

\subsection{Transfer Learning with Few Shots (\fs)}
\label{fs:results}
We now turn to answer \textbf{RQ3} concerning the effect of \emph{adding} few shots from the target language. Table~\ref{tab:rq3} shows the results when we \emph{continue} fine-tuning each of the models from Section~\ref{result:zero-shot} using \textbf{1\%} of the target language training set. For all target languages but English, this means we continue training using randomly sampled \textbf{17} examples from the target language, among which \textbf{8} are positive. As for English, we sample \textbf{9} examples, with \textbf{4} of them being positive. For sampling those additional examples, we fix the random seed across all languages. In additional (not shown) experiments with other seeds, we observe some variance in performance per model, which is consistent with a recent work on few shots learning~\cite{zhao2021closer}. However, we find that the overall observations are not affected by the different samples, thus we report results using a single sample seed. The baseline per language is on the diagonal of Table~\ref{tab:rq3}.

\begin{table}
  \caption{Effect of continued fine-tuning using 1\% of the target language training set. Values in () represent the percentage of difference between the results for this setup and corresponding \zs{} cell from Table~\ref{tab:rq1_zero}. $^{*}$ indicates significant difference from baseline on same target language.}
  \label{tab:rq3}
    \small
  \begin{tabular}{l|lllll}
    \toprule
     & \multicolumn{5}{c}{\bf target}\\
    \cline{2-6}
\multirow{2}{*}[1em]{\bf source} & \multicolumn{1}{c}{\bf ar} & \multicolumn{1}{c}{\bf bg}& \multicolumn{1}{c}{\bf en} & \multicolumn{1}{c}{\bf es} & \multicolumn{1}{c}{\bf tr}\\
\hline
  {\bf ar} & {\cellcolor [gray]{0.8}}49.5 & 35.9$^{*}$ (2) & 11.4$^{*}$ (-6) & 27.2$^{*}$ (2) & {\bf 44.1}$^{*}$ (-12)\\
  {\bf bg} & {\bf 55.1} (1.1K) & {\cellcolor [gray]{0.8}}{\bf 58.2} &  13.1 (-21)& 27.8$^{*}$ (276) & 36.5 (33)\\
  {\bf en} & 48.4 (1) & \underline{40.9}$^{*}$ (0)& {\cellcolor [gray]{0.8}}13.3 & \underline{30.5}$^{*}$ (11) & \underline{38.1} (-17) \\
  {\bf es} & 38.7 (12.8K) & 28.8$^{*}$ (330)& {\bf 14.5} (31)&{\cellcolor [gray]{0.8}}{\bf 54.0} & 31.5 (27)\\
  {\bf tr} & \underline{51.2} (316)& 37.1$^{*}$ (81)& \underline{13.8} (-18)& 25.6$^{*}$ (64)&{\cellcolor [gray]{0.8}}28.4 \\
\bottomrule
\end{tabular}
\end{table}

\begin{figure*}[htbp]
\centering
\includegraphics[scale=.25]{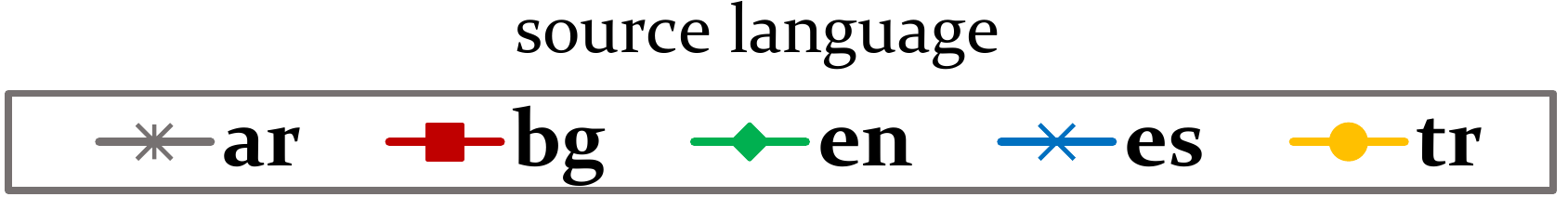}
\\[-2ex]
\subfloat[Target language is Arabic.]{\includegraphics[width=0.3\textwidth]{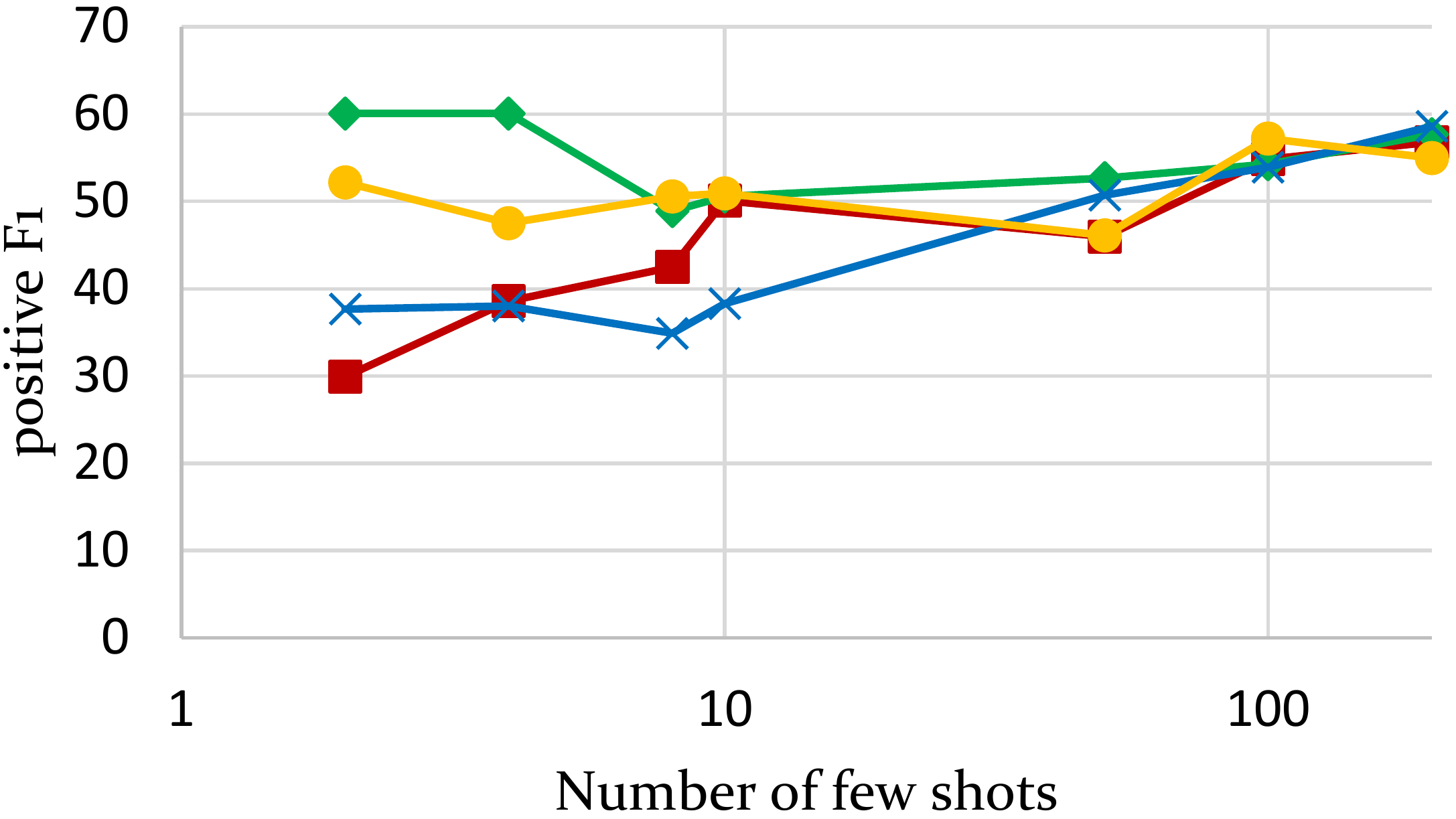}}\,
\subfloat[Target language is Bulgarian.]{\includegraphics[width=0.3\textwidth]{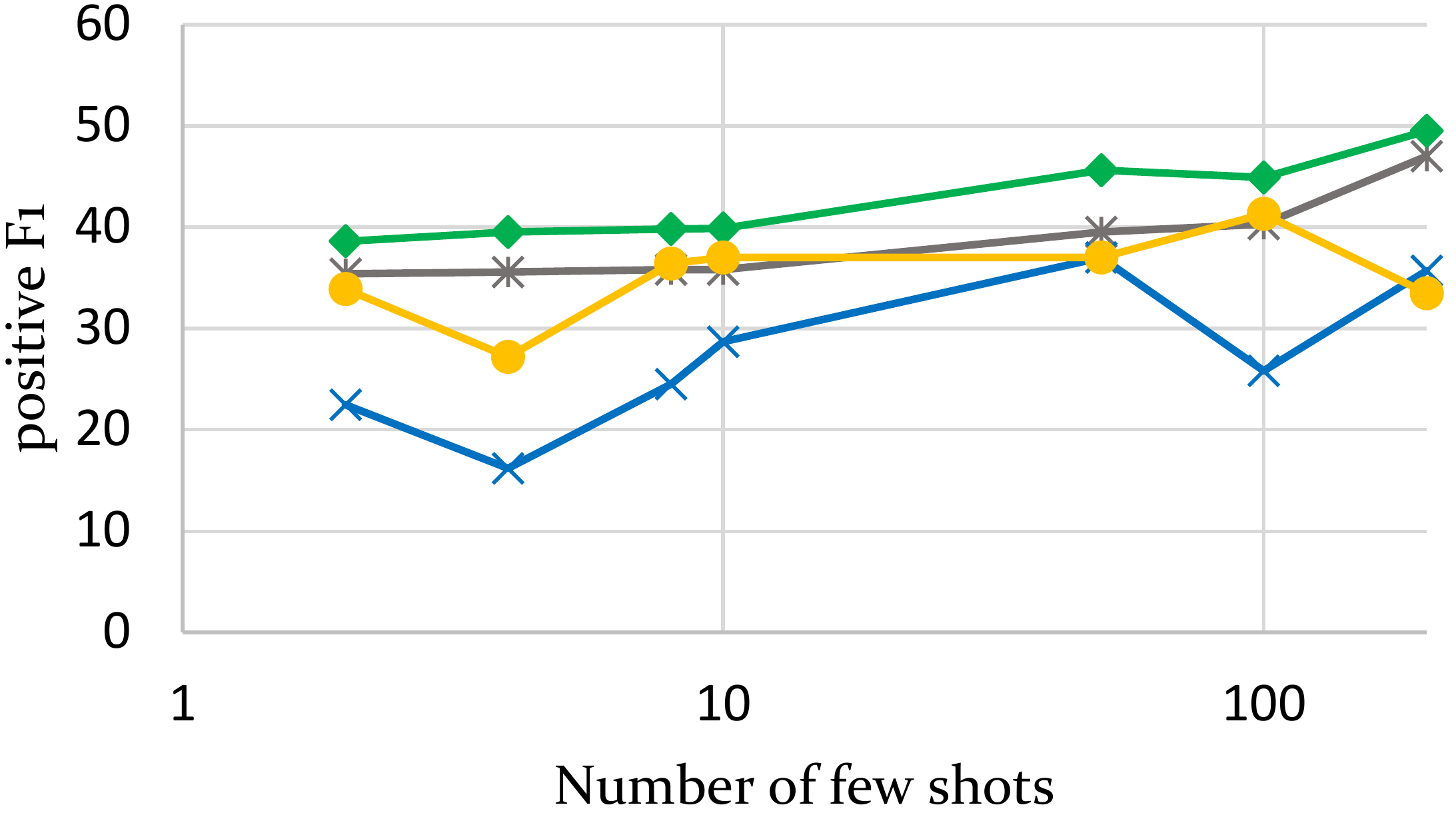}}\,
\subfloat[Target language is English.]{\includegraphics[width=0.3\textwidth]{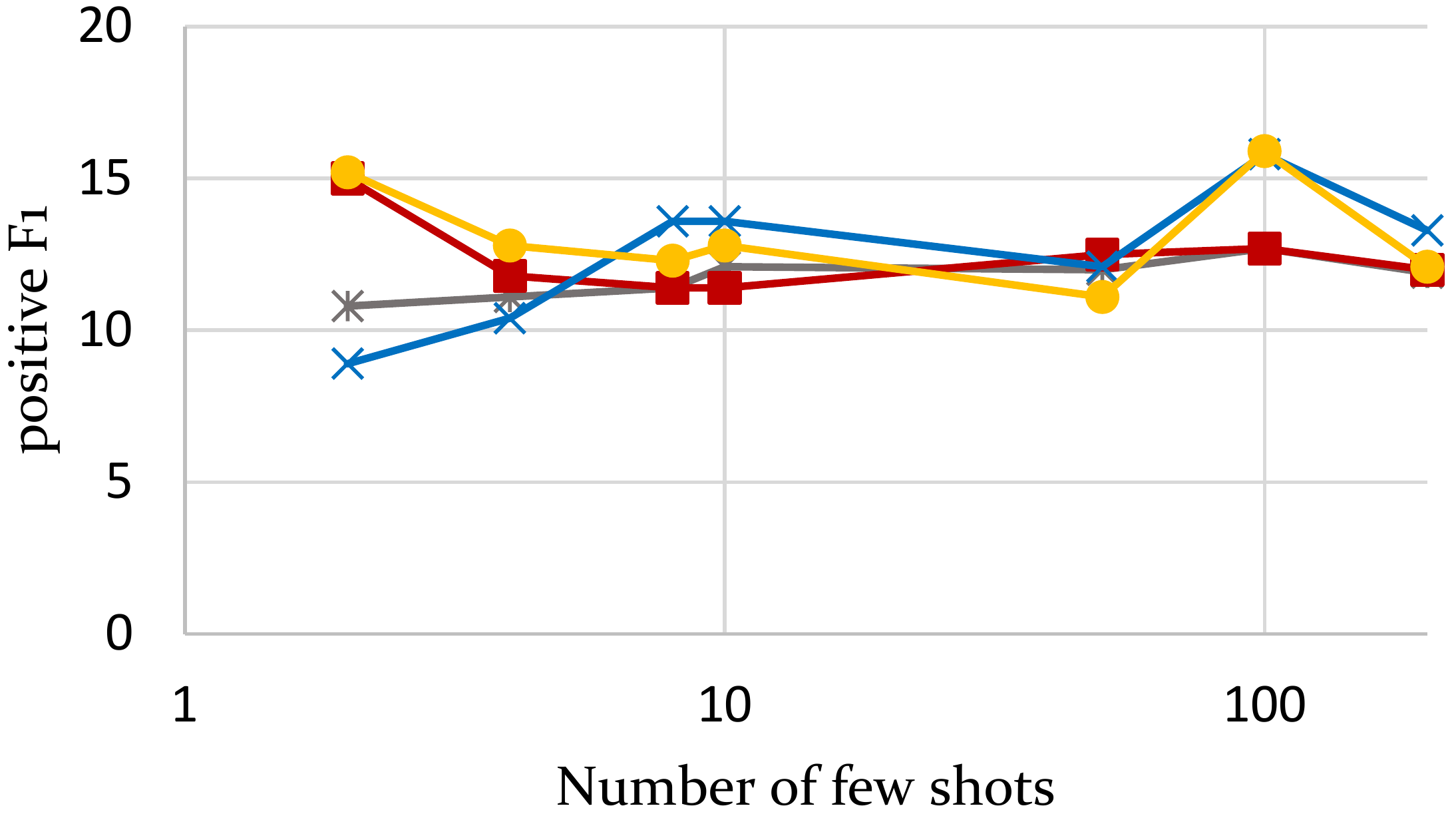}}
\\[-2ex]
\subfloat[Target language is Spanish.]{\includegraphics[width=0.3\textwidth]{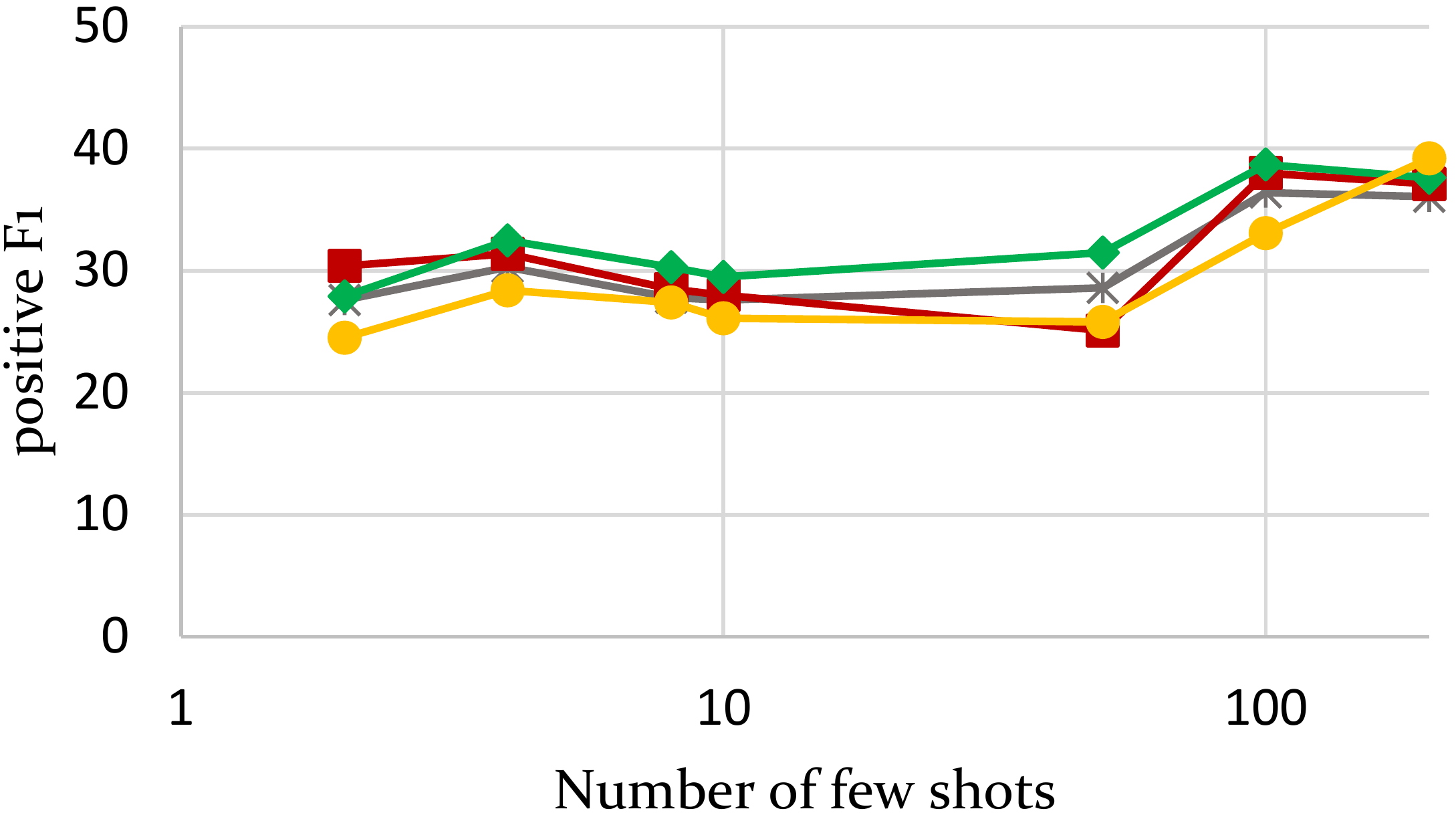}}\,
\subfloat[Target language is Turkish.]{\includegraphics[width=0.3\textwidth]{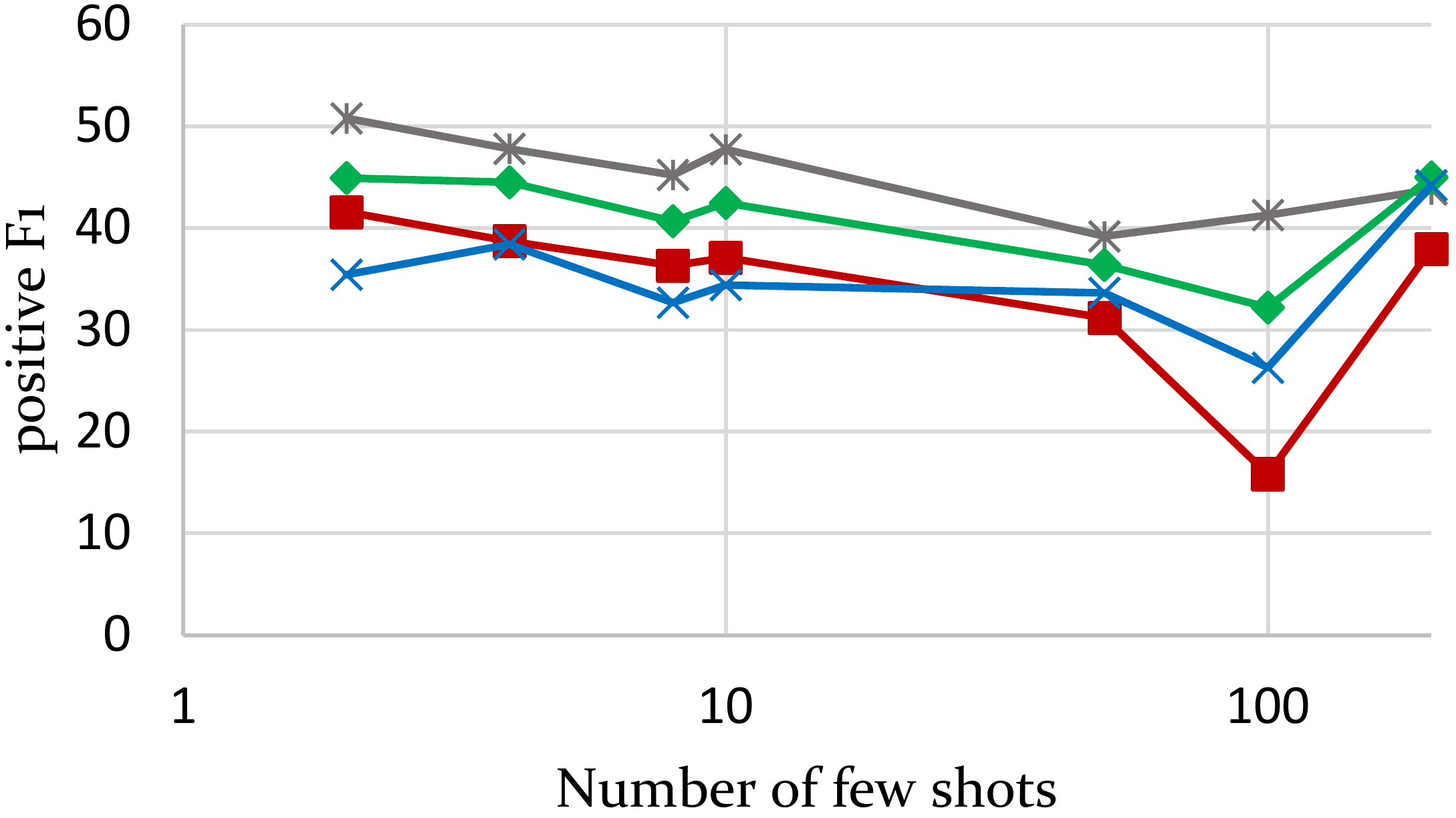}}
\caption{Effect of number of few shots on check-worthiness prediction. $x$-axis is in log scale.}
\label{fig:rq3_no_fewshots}
\end{figure*}

Compared to \zs, we observe better overall improvements. For the 20 language pairs, we find that few-shots learning improved over \zs{} in 14 cases with an average improvement of 15.8 points in $F_1$. This indicates that this setup is better than the case when we apply translation (maximum observed was 5.7 points).

The experiments further showed that adding as few as 17 Arabic examples to models, fine-tuned on each of the other languages except Spanish, resulted in comparable performance to the baseline, in which we train on the \emph{whole} Arabic training set. We also observe extreme improvements compared to pure \zs, except when English is the source language; for English, we observe that the performance is comparable. We anticipate this happened because Arabic is written in very different script (as opposed to remaining languages) making the addition of few target examples useful for \mbt{} to learn necessary structural and lexical information, which is consistent with observations made in a recent study~\cite{zhao2021closer} regarding the effect of writing scripts on \fs. These results highlight that with very small annotation effort, we can achieve as good performance as the case where much more annotations in the target language were collected. This is inline with recent studies on other text classification tasks~\cite{hedderich2020}. 

What happens when we change the number of the few shots? Figure~\ref{fig:rq3_no_fewshots} shows the effect of continuing the fine-tuning of \zs{} models with $k$ randomly sampled examples of the target language. We experiment with $k\in\{ 2,4,8,10,50,100,200\}$. The figure indicates several observations. First, for most target languages, an optimal setting of parameter $k$ is needed to get most benefit of few shots learning transfer, but we need to consider the corresponding annotation cost. Second, Arabic is generally the language that benefited the most from this setup, showing increasing performance gain with the increase of $k$, regardless of the source language. Third, for all target languages, except Bulgarian, we observe an interesting pattern, where at the addition of 200 shots from the target language, the achieved performance is very similar regardless of the source language. This indicates that the added 200 shots were enough to almost \emph{suppress} the effect of the source language. Fourth, for Spanish and Turkish, we observe a consistent pattern in performance across all source languages as we change the number of the few shots. Finally, there are clear source language winners in most cases, regardless of the number of few shots. English as a source language was almost always better for Arabic, Bulgarian, and Spanish, while Arabic as a source language was almost always better for Turkish. However, there was no clear winner for English as a target language, with $F_1$ limited between 10-15. This is somewhat expected, given that only 5\% of English test set is positive, making it very sensitive to mistakes in prediction.

\emph{In response to \textbf{RQ3}, \fs{} with the addition of as little as 1\% of the target language training set, has resulted in notable improvements over \zs{} for most language pairs. We also observe that effectiveness of \fs{} depends on finding an optimal setting for number of few shots for different language pairs, as adding more shots did not always result in improved performance.}   

\begin{figure*}[htbp]
\centering
\subfloat[Source language is Arabic]{\includegraphics[width=0.32\textwidth]{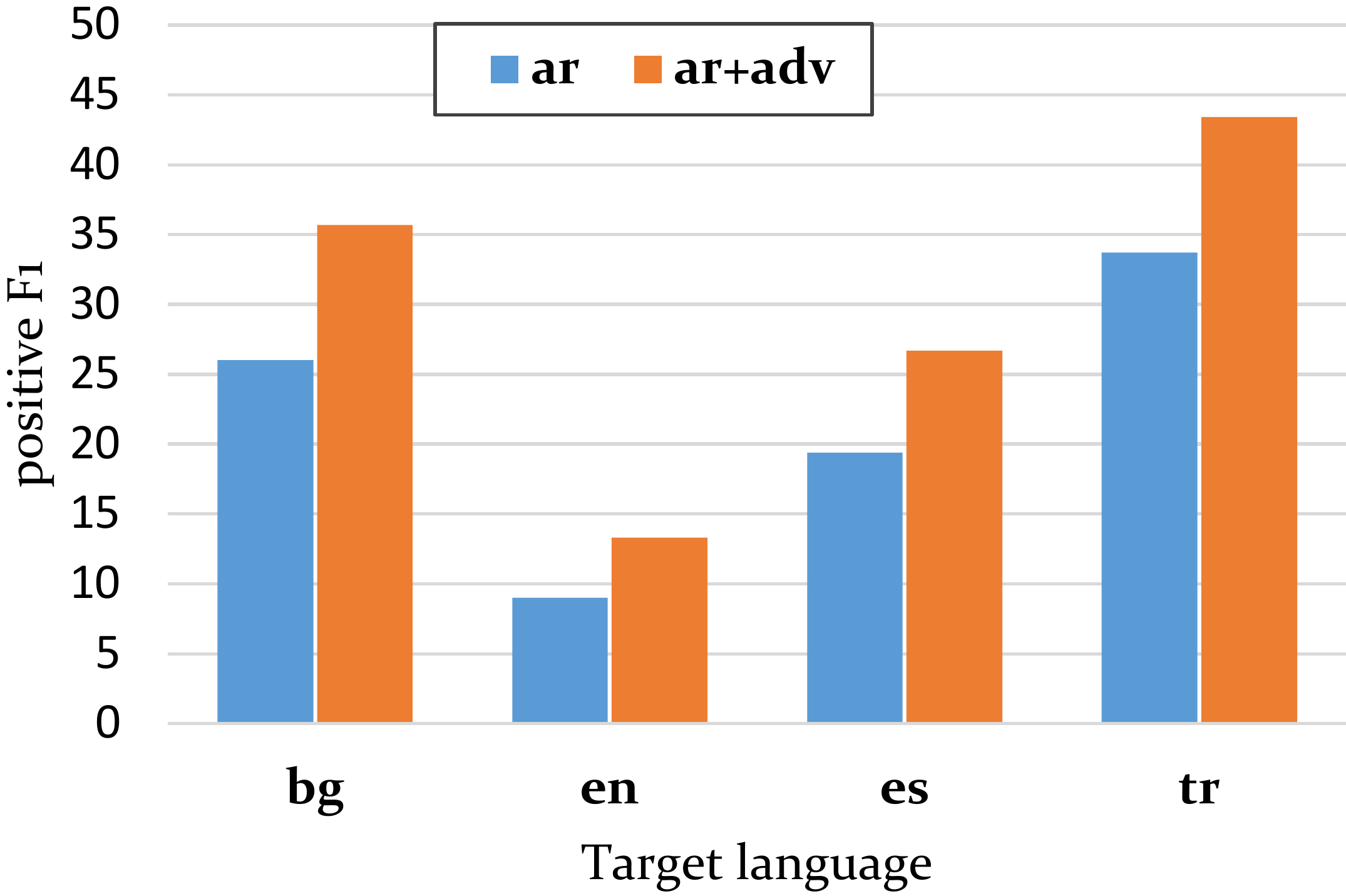}}\,
\subfloat[Source language is Bulgarian]{\includegraphics[width=0.32\textwidth]{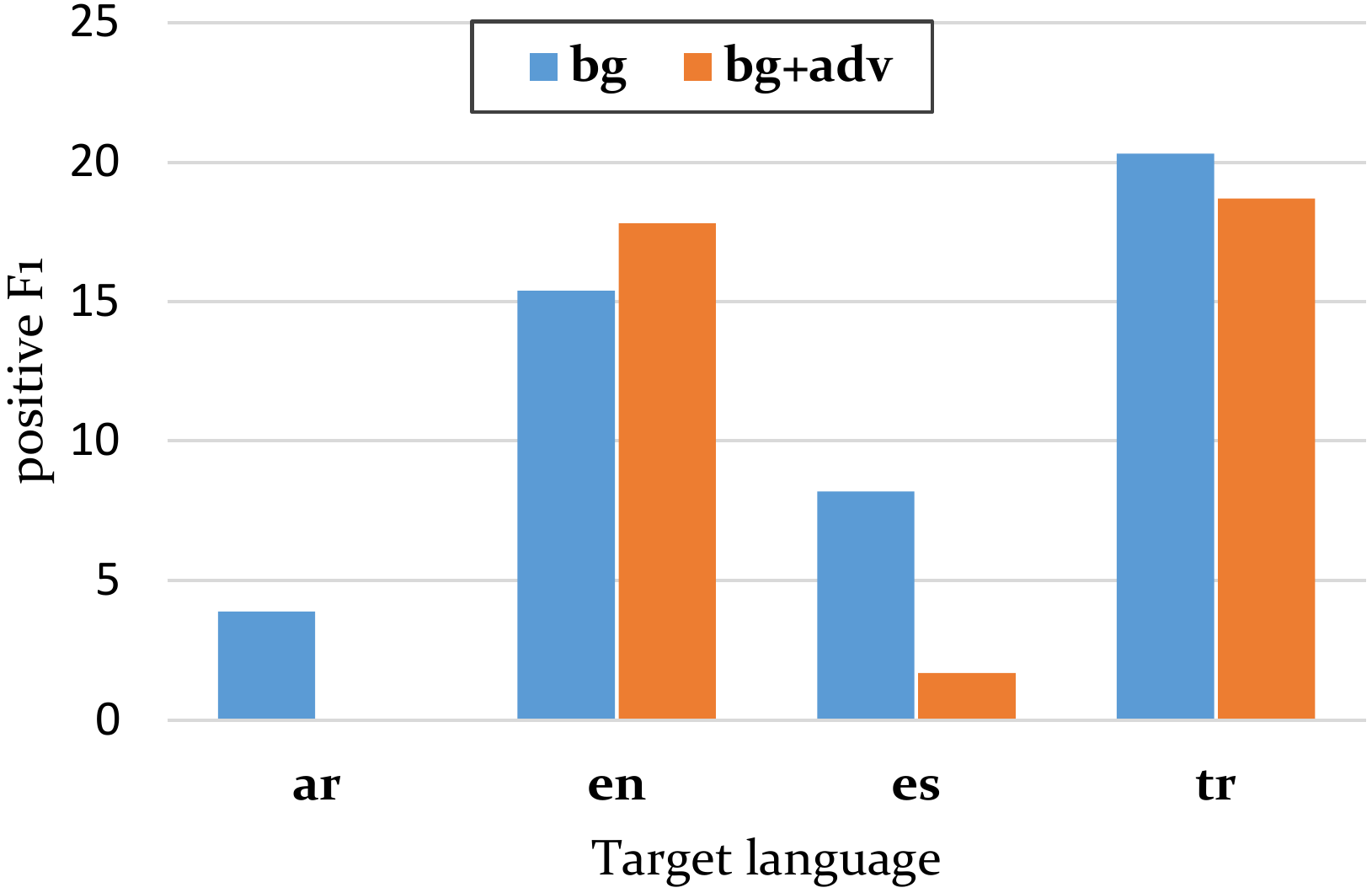}}\,
\subfloat[Source language is English]{\includegraphics[width=0.32\textwidth]{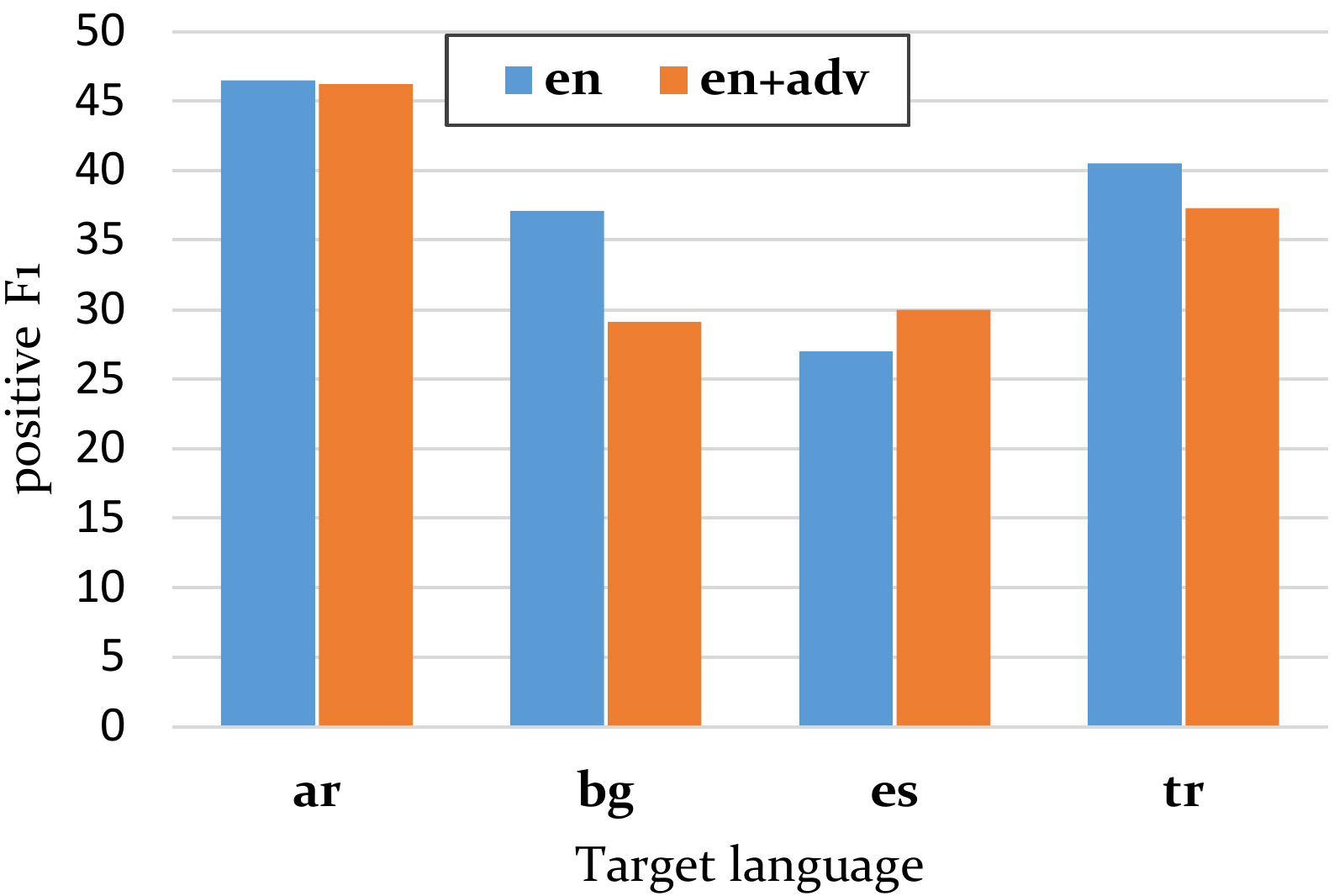}}
\caption{Effect of adversarial training on check-worthiness prediction. 
}
\label{fig:rq4}
\end{figure*}

\subsection{\zs{} with Adversarial Training (\advzs)}
\label{results:adv}
In this experiment, we address \textbf{RQ4}. We randomly sample \emph{unlabelled} examples from the target language training set and add them to the the training set of the source language to train the adversarial model explained in Section~\ref{sec:adv}. Figure~\ref{fig:rq4} shows the performance of this model in comparison to the \zs{} model. Note here that due to the size of the adversarial model and limited memory, we decrease the batch size to be 8 per model (including re-training the \zs{} model with this new batch size). We omit results for Spanish and Turkish source languages due to the results being close to zero.

The figures show conflicting observations. On one hand, \advzs{} was helpful when transferring from Arabic to other languages. On the other hand, we generally observe degradation in performance when transferring from Bulgarian or English to other languages. Moreover, none of the target languages except English consistently benefited from \advzs. We note here that this preliminary experiment was conducted using a single adversarial training approach for cross-lingual transfer. There might be more effective approaches for the classification task at hand, which we leave for future work.

\subsection{Multilingual \zs}
\begin{table}
  \caption{Performance of multilingual \zs{} model on the target language test sets. $^{*}$ indicates significant difference from baseline \mbt{}$_{target}$ on same test set.}
  \label{tab:rq5}
    \small
  \begin{tabular}{l|ccccc}
    \toprule
{\bf Model} &   {\bf ar} &   {\bf bg} &   {\bf en} &   {\bf es} &   {\bf tr} \\
\hline
{\bf \mbt{}$_{target}$} & \textbf{49.5} & \textbf{58.2} &  13.3 & \textbf{54.0} & 28.4 \\

{\bf \zs$_{best}$} & 47.7 & 41.0$^{*}$ & \textbf{16.8} & 27.6$^{*}$ & \textbf{50.3}$^{*}$ \\

{\bf \zs{}$_{all-target}$} & 24.4$^{*}$ & 26.8$^{*}$ & 13.5 & 25.6$^{*}$ & 34.0 \\
\bottomrule
\end{tabular}
\end{table}

\begin{figure}[htbp]
\centering
\subfloat[Fixed training data size\label{fig:rq5:fixed}]{\includegraphics[width=0.35\textwidth]{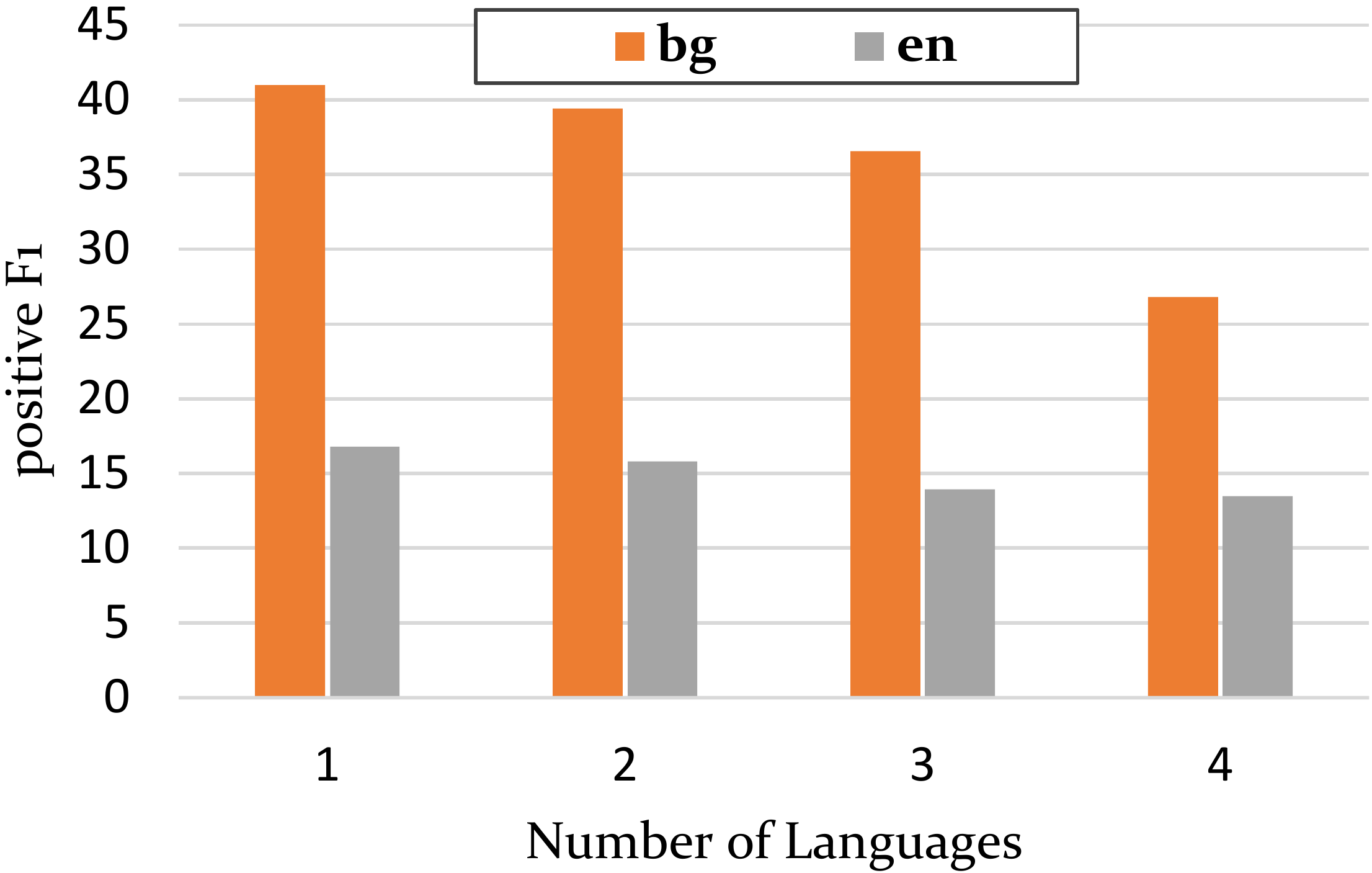}}
\\
\subfloat[Increasing training data size\label{fig:rq5:inc}]{\includegraphics[width=0.35\textwidth]{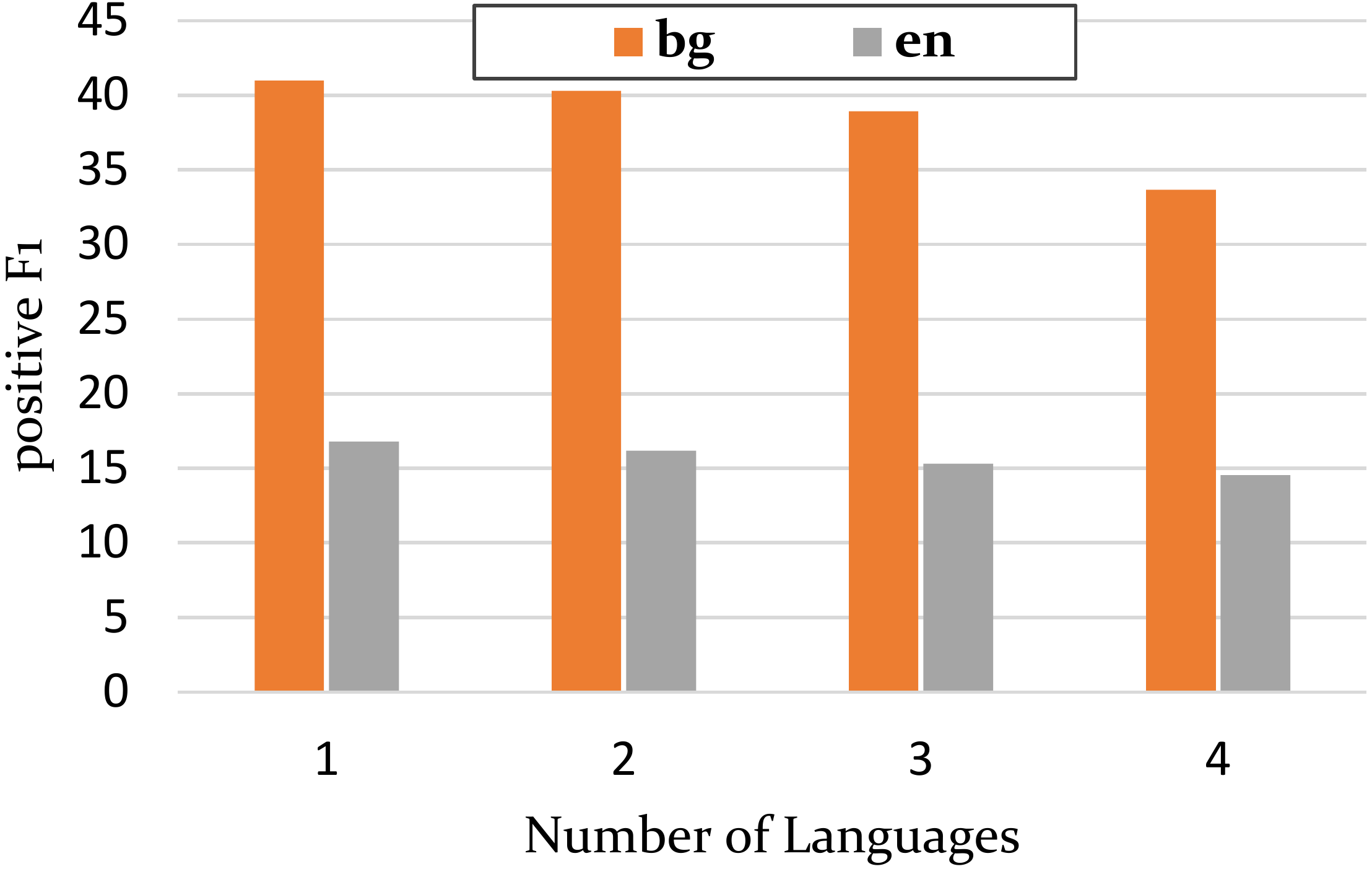}}
\caption{Effect of number of source languages used during fine-tuning on transfer performance to two target languages.}
\label{fig:rq5}
\end{figure}

We finally address \textbf{RQ5}: will \zs{} benefit from multilingual training? Differently from vanilla \zs{} (Section~\ref{result:zero-shot}) and typically used in literature, we fine-tune our check-worthiness prediction model over multilingual examples excluding the target language. For each language but the target language, we randomly sample examples with the same fixed class priors, ending up with 1,700 total examples across all four languages, with 300 positive examples and each language is equally represented in the training set. 
This is equal in total size and distribution to the training sets of the target languages, to ensure fair comparisons.

Table~\ref{tab:rq5} compares results of this model, denoted as \textbf{\zs{}$_{all-target}$}, with two baselines: (1) \textbf{\mbt{}$_{target}$}, in which we fine-tune the \mbt{} model over the training set of the target language, and (2) \textbf{\zs$_{best}$}, which is the best per-target-language performing \zs{} model in Table~\ref{tab:rq1_zero}, which is fine-tuned with \emph{one} source language. We observe that for 3 out of the 5 languages, unified language in training and test sets (i.e., \mbt{}$_{target}$) yields the best performance. Surprisingly, this experiment shows that multilinguality in training did not actually help the model achieve better performance compared to the vanilla \zs. We argue that this is due to the \emph{curse of multilinguality} that was observed on multilingual transformer models due to limited model capacity dedicated to each language during pre-training, causing degraded transfer ability~\cite{conneau2020unsupervised}. We believe a similar issue is emerging during fine-tuning using multiple languages; with each language being represented by small number of examples, the transfer ability of the model, compared to pure \zs{} setup, is negatively affected. 

We further support our claim by running an experiment in which we incrementally increase the number of source languages (excluding the target language) during fine-tuning, and report the transfer performance on a low resource (\textbf{bg}) and a high resource (\textbf{en}) target languages. As with \textbf{\zs{}$_{all-target}$}, we fix the total training examples to 1.7k and keep each source language equally represented in the training sample. For the case of 1, 2, or 3 source languages, we try all combinations; thus, for those cases, we end up with multiple performance values depending on the combination of source language(s) used during fine-tuning; we report the maximum observed scores for (\textbf{bg}) and (\textbf{en}) in Figure~\ref{fig:rq5:fixed}. The figure shows the transfer performance degrades for both languages as we increase the number of languages on which \mbt{} is fine-tuned. This is consistent with the phenomenon observed by~\citet{conneau2020unsupervised} for transformer models during pre-training. 

A question arises on whether this effect can be mitigated by increasing the training set size as we increase the number of languages. 
We repeated the previous experiment, but increased training set size to be 1,700 per source language. Figure~\ref{fig:rq5:inc} shows, again, degradation in performance but not as severe. \emph{Overall, experiments indicate that increasing the number of source languages (i.e., introducing multilinguality in training set) is not as effective for our task.}     

\subsection{Comparison to State of the Art Models}
Answering \textbf{RQ6}, we would like now to see how the previous setups compare to baselines that we hypothesize are the state of the art \emph{on the given test set}. We experiment with the following baselines:
\begin{enumerate}
    \item \textbf{\mbt{}$_{target}$}, as described earlier.
    \item \textbf{\monbt{}$_{target}$}, in which we fine-tune a monolingual pre-trained BERT model per target language (e.g., \abt{} for Arabic) using the training set of the target language. The models we test are those shown in Table~\ref{tab:models}. Such setup is expected to be effective, since the monolingual model is pre-trained on a large corpora in the target language.
    \item \textbf{CT!2021$_{best}$}, which is the model with best reported performance per target language in CheckThat! 2021 lab~\cite{clef-checkthat:2021:task1}.\footnote{We note that the systems were originally evaluated as ranking systems, however, to facilitate our comparison, we re-evaluate the runs as classification systems. To that purpose, we use the prediction of check-worthiness score per tweet as it appeared in the original run files we acquired from the teams. This is possible since all teams report using classification models to solve the task and report probability of prediction (between 0 and 1) as the ranking score. We assume any tweet with score above 0.5 to be predicted as check-worthy claim.}
\end{enumerate}

Table~\ref{tab:rq6} compares the performance of the above baselines with the best performing models per setup from those presented above, where none or minimal labeled examples in the target language were considered in fine-tuning. The comparison yields a clear observation; for three languages, \textbf{ar}, \textbf{en}, and \textbf{tr}, at least one of the \zs{} setups had comparable or insignificantly-different performance to both of the \bt{}-based baselines trained on the training set of the target language. More notably, for two languages, \textbf{en} and \textbf{tr}, translation yields performances gains that are significantly different from those two baselines. Moreover, our best \trzs{} model outperforms all baselines on \textbf{en}. Overall, the comparisons indicate that it is possible to train an effective cross-lingual check-worthiness model for at least three languages without the need for any training examples in the target language.

\begin{table}
  \caption{Comparison of performance of best \zs{} setup for each target language and state-of-the-art models. $^{*}$, $\dagger$ indicate significant difference from \mbt{}$_{target}$, \monbt{}$_{target}$ on same test set.}
  \label{tab:rq6}
    \small
  \begin{tabular}{l|ccccc}
    \toprule
{\bf Model} &   {\bf ar} &   {\bf bg} &   {\bf en} &   {\bf es} &   {\bf tr} \\
\hline
{\bf CT!2021$_{best}$} & {\bf 60.0} & {\bf 63.9} & 15.3 & 27.0 & {\bf 53.9} \\
{\bf \mbt{}$_{target}$} & 49.5 &  \underline{58.2} & 13.3 & {\bf 54.0} & 28.4 \\
{\bf \monbt{}$_{target}$} & \underline{56.3} & 57.8 & 14.2 & \underline{49.0} & 37.2 \\
\hline
{\bf \zs$_{best}$} & 47.7 & 41.0$^{*,\dagger}$ & \underline{16.8} & 27.6$^{*,\dagger}$ & 50.3$^{*}$ \\
{\bf \trzs$_{best}$} & 45.7 & 36.3$^{*,\dagger}$ & {\bf 17.9}$^{*}$ & 30.0$^{*,\dagger}$ & 47.0 \\
{\bf \tezs$_{best}$} & 52.0 & 40.9$^{*,\dagger}$ & 14.1 & 27.9$^{*,\dagger}$ & \underline{51.9}$^{*}$ \\
{\bf \fs$_{best}$} & 55.1 & 40.9$^{*,\dagger}$ & 14.5 & 30.5$^{*,\dagger}$ & 44.1$^{*}$ \\
\bottomrule
\end{tabular}
\end{table}

However, we note that for \textbf{bg} and \textbf{es}, the proposed models consistently under-perform compared to \bt-based baselines. 
We believe that for \textbf{Bulgarian} compared to \mbt{}$_{target}$, this is due to the initial pre-training of \mbt{}, where Bulgarian (among other low-resource languages) was greatly under-represented compared to English and other higher-resource languages.\footnote{\url{https://github.com/google-research/bert/blob/master/multilingual.md}} 
We are aware that these results on Bulgarian are not consistent with those in a relevant recent study~\cite{panda2021detecting}. In that study, a multitask learning approach was adopted using \mbt{} for cross-lingual or few-shot transfer learning on seven joint tasks including check-worthiness. Due to the evaluation being done as a weighted score over all tasks, that study did not show clearly how the cross-lingual \mbt{} model performed for check-worthiness task specifically, and even showed that, overall, the proposed model is as a good as monolingual models trained on the full training set. This further demonstrates the importance of our work for check-worthiness estimation, where we managed to identify that for Bulgarian, cross-lingual transfer learning is not effective. 

As for \textbf{Spanish}, we note that the annotation strategy for this language is slightly different, which might have caused a difference in the criteria of check-worthiness compared to the other languages. 
That is, the problem is no longer transfer between languages, but also between slightly different classification tasks. In Section~\ref{result:transtest} we also observed potential generalization limitation of the models between topics from the training and test sets. This raises a ``yet to be answered'' question on the accuracy of the currently available claim check-worthiness definitions in terms of how well they match true user needs. For example, should the topic be considered as part of estimating check-worthiness or not?  We leave answering this question and studying sensitivity of cross-lingual transfer across datasets with different definitions of check-worthiness or different topics to future work.

The table also shows that systems from CheckThat! 2021 achieved the best performance for \textbf{ar}, \textbf{bg}, and \textbf{tr}. We note that this is a bit unfair comparison, since those systems were trained on much larger training sets (recall that we under-sample the training sets from CT!2021 in all our experiments (Table~\ref{tab:datasets})). Even with that disadvantage to our models, the models still show comparable performance to those from CT!2021 on three languages (\textbf{en}, \textbf{es}, and \textbf{tr}). This demonstrates the strong cross-lingual transfer ability of \mbt{} for our problem, especially when translation is employed to unify the source and target languages. \emph{A more important observation here is that with as few as 1.7k labelled examples, which, in our experience, are not very difficult to acquire for the check-worthiness prediction task, the cross-lingual models achieve comparable performance to state-of-the-art models from the CheckThat! lab on this dataset for those languages.}

\section{Conclusions and Future Work}
\label{conc}
In this study, we aimed to investigate and identify optimal setups to facilitate check-worthiness estimation over multilingual streams. Our work is motivated by the current dire need to provide multilingual support for the problem given the scale of propagating misinformation and the parallel efforts being put to verify the same claims across languages. Our in-depth experiments showed that cross-lingual transfer models result in comparable performance to the monolingual models trained on examples in the target language for at least three languages, Arabic, English and Turkish. Moreover, multilinguality during fine-tuning negatively affected the model transfer performance. We also showed that our proposed models are as effective as the state of the art models on English, Spanish and Turkish. Our experiments did not show much benefit from employing adversarial training to decrease model over-fitting to source language. Finally, the addition of few shots showed to be generally helpful for three target languages (Arabic, Bulgarian and Spanish) compared to zero shot learning transfer setups.

For future work, we plan to investigate more advanced adversarial training setups. We are also interested in investigating the performance of other multilingual transformer models (e.g., XLM-R). Another natural extension to the work is to experiment with more languages and even out-of-domain datasets.

\begin{acks}
This work was made possible by NPRP grant \# NPRP 11S-1204-170060 from the Qatar National Research Fund (a member of Qatar Foundation). The statements made herein are solely the responsibility of the authors.
\end{acks}

\bibliographystyle{ACM-Reference-Format}
\bibliography{cw_ref.bib}


\begin{thebibliography}{49}


\ifx \showCODEN    \undefined \def \showCODEN     #1{\unskip}     \fi
\ifx \showDOI      \undefined \def \showDOI       #1{#1}\fi
\ifx \showISBNx    \undefined \def \showISBNx     #1{\unskip}     \fi
\ifx \showISBNxiii \undefined \def \showISBNxiii  #1{\unskip}     \fi
\ifx \showISSN     \undefined \def \showISSN      #1{\unskip}     \fi
\ifx \showLCCN     \undefined \def \showLCCN      #1{\unskip}     \fi
\ifx \shownote     \undefined \def \shownote      #1{#1}          \fi
\ifx \showarticletitle \undefined \def \showarticletitle #1{#1}   \fi
\ifx \showURL      \undefined \def \showURL       {\relax}        \fi
\providecommand\bibfield[2]{#2}
\providecommand\bibinfo[2]{#2}
\providecommand\natexlab[1]{#1}
\providecommand\showeprint[2][]{arXiv:#2}

\bibitem[cle(2020)]%
        {clef2020-workingnotes}
 \bibinfo{year}{2020}\natexlab{}.
\newblock \showarticletitle{CLEF 2020 Working Notes}
  \emph{(\bibinfo{series}{{CEUR} Workshop Proceedings})},
  \bibfield{editor}{\bibinfo{person}{Linda Cappellato},
  \bibinfo{person}{Carsten Eickhoff}, \bibinfo{person}{Nicola Ferro}, {and}
  \bibinfo{person}{Aurélie Névéol}} (Eds.).
\newblock
\showISSN{1613-0073}


\bibitem[cle(2021)]%
        {clef2021-workingnotes}
 \bibinfo{year}{2021}\natexlab{}.
\newblock \showarticletitle{{CLEF} 2021 Working Notes. {W}orking Notes of
  {CLEF} 2021--Conference and Labs of the Evaluation Forum},
  \bibfield{editor}{\bibinfo{person}{Guglielmo Faggioli},
  \bibinfo{person}{Nicola Ferro}, \bibinfo{person}{Alexis Joly},
  \bibinfo{person}{Maria Maistro}, {and} \bibinfo{person}{Florina Piroi}}
  (Eds.).
\newblock


\bibitem[Alam et~al\mbox{.}(2021a)]%
        {Alam_2021}
\bibfield{author}{\bibinfo{person}{Firoj Alam}, \bibinfo{person}{Fahim Dalvi},
  \bibinfo{person}{Shaden Shaar}, \bibinfo{person}{Nadir Durrani},
  \bibinfo{person}{Hamdy Mubarak}, \bibinfo{person}{Alex Nikolov},
  \bibinfo{person}{Giovanni Da~San~Martino}, \bibinfo{person}{Ahmed Abdelali},
  \bibinfo{person}{Hassan Sajjad}, \bibinfo{person}{Kareem Darwish}, {and}
  \bibinfo{person}{Preslav Nakov}.} \bibinfo{year}{2021}\natexlab{a}.
\newblock \showarticletitle{Fighting the COVID-19 Infodemic in Social Media: A
  Holistic Perspective and a Call to Arms}.
\newblock \bibinfo{journal}{\emph{Proceedings of the International AAAI
  Conference on Web and Social Media}} \bibinfo{volume}{15},
  \bibinfo{number}{1} (\bibinfo{year}{2021}), \bibinfo{pages}{913--922}.
\newblock


\bibitem[Alam et~al\mbox{.}(2021b)]%
        {alam2021-fighting-covid}
\bibfield{author}{\bibinfo{person}{Firoj Alam}, \bibinfo{person}{Shaden Shaar},
  \bibinfo{person}{Fahim Dalvi}, \bibinfo{person}{Hassan Sajjad},
  \bibinfo{person}{Alex Nikolov}, \bibinfo{person}{Hamdy Mubarak},
  \bibinfo{person}{Giovanni Da~San~Martino}, \bibinfo{person}{Ahmed Abdelali},
  \bibinfo{person}{Nadir Durrani}, \bibinfo{person}{Kareem Darwish},
  \bibinfo{person}{Abdulaziz Al-Homaid}, \bibinfo{person}{Wajdi Zaghouani},
  \bibinfo{person}{Tommaso Caselli}, \bibinfo{person}{Gijs Danoe},
  \bibinfo{person}{Friso Stolk}, \bibinfo{person}{Britt Bruntink}, {and}
  \bibinfo{person}{Preslav Nakov}.} \bibinfo{year}{2021}\natexlab{b}.
\newblock \showarticletitle{Fighting the {COVID}-19 Infodemic: Modeling the
  Perspective of Journalists, Fact-Checkers, Social Media Platforms, Policy
  Makers, and the Society}. In \bibinfo{booktitle}{\emph{Findings of the
  Association for Computational Linguistics: EMNLP 2021}}.
  \bibinfo{publisher}{Association for Computational Linguistics},
  \bibinfo{address}{Punta Cana, Dominican Republic}, \bibinfo{pages}{611--649}.
\newblock
\urldef\tempurl%
\url{https://aclanthology.org/2021.findings-emnlp.56}
\showURL{%
\tempurl}


\bibitem[Alhindi et~al\mbox{.}(2021)]%
        {alhindi2021fact}
\bibfield{author}{\bibinfo{person}{Tariq Alhindi}, \bibinfo{person}{Brennan
  McManus}, {and} \bibinfo{person}{Smaranda Muresan}.}
  \bibinfo{year}{2021}\natexlab{}.
\newblock \showarticletitle{What to Fact-Check: Guiding Check-Worthy
  Information Detection in News Articles through Argumentative Discourse
  Structure}. In \bibinfo{booktitle}{\emph{Proceedings of the 22nd Annual
  Meeting of the Special Interest Group on Discourse and Dialogue}}.
  \bibinfo{pages}{380--391}.
\newblock


\bibitem[Antoun et~al\mbox{.}(2020)]%
        {antoun2020arabert}
\bibfield{author}{\bibinfo{person}{Wissam Antoun}, \bibinfo{person}{Fady Baly},
  {and} \bibinfo{person}{Hazem Hajj}.} \bibinfo{year}{2020}\natexlab{}.
\newblock \showarticletitle{AraBERT: Transformer-based Model for Arabic
  Language Understanding}. In \bibinfo{booktitle}{\emph{LREC 2020 Workshop
  Language Resources and Evaluation Conference}}. \bibinfo{pages}{9}.
\newblock


\bibitem[Arkhipov et~al\mbox{.}(2019)]%
        {arkhipov-etal-2019-tuning}
\bibfield{author}{\bibinfo{person}{Mikhail Arkhipov}, \bibinfo{person}{Maria
  Trofimova}, \bibinfo{person}{Yuri Kuratov}, {and} \bibinfo{person}{Alexey
  Sorokin}.} \bibinfo{year}{2019}\natexlab{}.
\newblock \showarticletitle{Tuning Multilingual Transformers for
  Language-Specific Named Entity Recognition}. In
  \bibinfo{booktitle}{\emph{Proceedings of the 7th Workshop on Balto-Slavic
  Natural Language Processing}}. \bibinfo{publisher}{Association for
  Computational Linguistics}, \bibinfo{address}{Florence, Italy},
  \bibinfo{pages}{89--93}.
\newblock
\urldef\tempurl%
\url{https://doi.org/10.18653/v1/W19-3712}
\showDOI{\tempurl}


\bibitem[Arnold(2020)]%
        {fullfact:coof}
\bibfield{author}{\bibinfo{person}{Phoebe Arnold}.}
  \bibinfo{year}{2020}\natexlab{}.
\newblock \bibinfo{booktitle}{\emph{The challenges of online fact checking}}.
\newblock \bibinfo{type}{{T}echnical {R}eport}. \bibinfo{institution}{Full
  Fact}.
\newblock


\bibitem[Arslan et~al\mbox{.}(2020)]%
        {arslan2020benchmark}
\bibfield{author}{\bibinfo{person}{Fatma Arslan}, \bibinfo{person}{Naeemul
  Hassan}, \bibinfo{person}{Chengkai Li}, {and} \bibinfo{person}{Mark
  Tremayne}.} \bibinfo{year}{2020}\natexlab{}.
\newblock \showarticletitle{A benchmark dataset of check-worthy factual
  claims}. In \bibinfo{booktitle}{\emph{Proceedings of the International AAAI
  Conference on Web and Social Media}}, Vol.~\bibinfo{volume}{14}.
  \bibinfo{pages}{821--829}.
\newblock


\bibitem[Atanasova et~al\mbox{.}(2018)]%
        {clef-checkthat-T1:2018}
\bibfield{author}{\bibinfo{person}{Pepa Atanasova}, \bibinfo{person}{Lluis
  Marquez}, \bibinfo{person}{Alberto Barr\'{o}n-Cede{\~n}o},
  \bibinfo{person}{Tamer Elsayed}, \bibinfo{person}{Reem Suwaileh},
  \bibinfo{person}{Wajdi Zaghouani}, \bibinfo{person}{Spas Kyuchukov},
  \bibinfo{person}{Giovanni Da~San~Martino}, {and} \bibinfo{person}{Preslav
  Nakov}.} \bibinfo{year}{2018}\natexlab{}.
\newblock \showarticletitle{Overview of the {CLEF-2018 CheckThat!} Lab on
  Automatic Identification and Verification of Political Claims. {T}ask 1:
  Check-Worthiness} \emph{(\bibinfo{series}{{CEUR} Workshop Proceedings})},
  \bibfield{editor}{\bibinfo{person}{Linda Cappellato}, \bibinfo{person}{Nicola
  Ferro}, \bibinfo{person}{Jian-Yun Nie}, {and} \bibinfo{person}{Laure
  Soulier}} (Eds.).
\newblock


\bibitem[Atanasova et~al\mbox{.}(2019)]%
        {clef-checkthat-T1:2019}
\bibfield{author}{\bibinfo{person}{Pepa Atanasova}, \bibinfo{person}{Preslav
  Nakov}, \bibinfo{person}{Georgi Karadzhov}, \bibinfo{person}{Mitra
  Mohtarami}, {and} \bibinfo{person}{Giovanni Da~San~Martino}.}
  \bibinfo{year}{2019}\natexlab{}.
\newblock \showarticletitle{{Overview of the CLEF-2019 CheckThat!} Lab on
  Automatic Identification and Verification of Claims. {T}ask 1:
  Check-Worthiness} \emph{(\bibinfo{series}{{CEUR} Workshop Proceedings})},
  \bibfield{editor}{\bibinfo{person}{Linda Cappellato}, \bibinfo{person}{Nicola
  Ferro}, \bibinfo{person}{{David E.} Losada}, {and} \bibinfo{person}{Henning
  M{\"u}ller}} (Eds.).
\newblock


\bibitem[{Baris Schlicht} et~al\mbox{.}(2021)]%
        {clef-checkthat:2021:task1:Schlicht2021}
\bibfield{author}{\bibinfo{person}{Ipek {Baris Schlicht}},
  \bibinfo{person}{{Angel Felipe} {Magnossão de Paula}}, {and}
  \bibinfo{person}{Paolo Rosso}.} \bibinfo{year}{2021}\natexlab{}.
\newblock \showarticletitle{{UPV} at {CheckThat! 2021}: Mitigating Cultural
  Differences for Identifying Multilingual Check-worthy Claims}, See
  \citeN{clef2021-workingnotes}.
\newblock


\bibitem[Beltrán et~al\mbox{.}(2021)]%
        {beltran2021claimhunter}
\bibfield{author}{\bibinfo{person}{Javier Beltrán}, \bibinfo{person}{Rubén
  Míguez}, {and} \bibinfo{person}{Irene Larraz}.}
  \bibinfo{year}{2021}\natexlab{}.
\newblock \showarticletitle{ClaimHunter: An unattended tool for automated claim
  detection on Twitter}. In \bibinfo{booktitle}{\emph{Proceedings of the 1st
  International Workshop on Knowledge Graphs for Online Discourse Analysis}}
  \emph{(\bibinfo{series}{KnOD 2021})}.
\newblock


\bibitem[Brennen et~al\mbox{.}(2020)]%
        {brennen2020}
\bibfield{author}{\bibinfo{person}{J.~Scott Brennen}, \bibinfo{person}{Felix~M.
  Simon}, \bibinfo{person}{Philip~N. Howard}, {and}
  \bibinfo{person}{Rasmus~Kleis Nielsen}.} \bibinfo{year}{2020}\natexlab{}.
\newblock \bibinfo{booktitle}{\emph{Types, sources, and claims of COVID-19
  misinformation}}.
\newblock Reuters Institute for the Study of Journalism, University of Oxford.
\newblock


\bibitem[Cañete et~al\mbox{.}(2020)]%
        {CaneteCFP2020}
\bibfield{author}{\bibinfo{person}{José Cañete}, \bibinfo{person}{Gabriel
  Chaperon}, \bibinfo{person}{Rodrigo Fuentes}, \bibinfo{person}{Jou-Hui Ho},
  \bibinfo{person}{Hojin Kang}, {and} \bibinfo{person}{Jorge Pérez}.}
  \bibinfo{year}{2020}\natexlab{}.
\newblock \showarticletitle{Spanish Pre-Trained BERT Model and Evaluation
  Data}. In \bibinfo{booktitle}{\emph{PML4DC at ICLR 2020}}.
\newblock


\bibitem[Chen et~al\mbox{.}(2019)]%
        {chen2019multi}
\bibfield{author}{\bibinfo{person}{Xilun Chen}, \bibinfo{person}{Ahmed
  Hassan~Awadallah}, \bibinfo{person}{Hany Hassan}, \bibinfo{person}{Wei Wang},
  {and} \bibinfo{person}{Claire Cardie}.} \bibinfo{year}{2019}\natexlab{}.
\newblock \showarticletitle{Multi-Source Cross-Lingual Model Transfer: Learning
  What to Share}. In \bibinfo{booktitle}{\emph{Proceedings of the 57th
  Conference of the Association for Computational Linguistics (ACL)}}.
\newblock


\bibitem[Conneau et~al\mbox{.}(2020)]%
        {conneau2020unsupervised}
\bibfield{author}{\bibinfo{person}{Alexis Conneau}, \bibinfo{person}{Kartikay
  Khandelwal}, \bibinfo{person}{Naman Goyal}, \bibinfo{person}{Vishrav
  Chaudhary}, \bibinfo{person}{Guillaume Wenzek}, \bibinfo{person}{Francisco
  Guzm{\'a}n}, \bibinfo{person}{{\'E}douard Grave}, \bibinfo{person}{Myle Ott},
  \bibinfo{person}{Luke Zettlemoyer}, {and} \bibinfo{person}{Veselin
  Stoyanov}.} \bibinfo{year}{2020}\natexlab{}.
\newblock \showarticletitle{Unsupervised Cross-lingual Representation Learning
  at Scale}. In \bibinfo{booktitle}{\emph{Proceedings of the 58th Annual
  Meeting of the Association for Computational Linguistics}}.
  \bibinfo{pages}{8440--8451}.
\newblock


\bibitem[Devlin et~al\mbox{.}(2019)]%
        {devlin2019}
\bibfield{author}{\bibinfo{person}{Jacob Devlin}, \bibinfo{person}{Ming-Wei
  Chang}, \bibinfo{person}{Kenton Lee}, {and} \bibinfo{person}{Kristina
  Toutanova}.} \bibinfo{year}{2019}\natexlab{}.
\newblock \showarticletitle{BERT: Pre-training of Deep Bidirectional
  Transformers for Language Understanding}. In
  \bibinfo{booktitle}{\emph{Proceedings of the 2019 Conference of the North
  American Chapter of the Association for Computational Linguistics: Human
  Language Technologies, Volume 1 (Long and Short Papers)}}.
  \bibinfo{pages}{4171--4186}.
\newblock


\bibitem[Dong et~al\mbox{.}(2020)]%
        {dong2020leveraging}
\bibfield{author}{\bibinfo{person}{Xin Dong}, \bibinfo{person}{Yaxin Zhu},
  \bibinfo{person}{Yupeng Zhang}, \bibinfo{person}{Zuohui Fu},
  \bibinfo{person}{Dongkuan Xu}, \bibinfo{person}{Sen Yang}, {and}
  \bibinfo{person}{Gerard De~Melo}.} \bibinfo{year}{2020}\natexlab{}.
\newblock \showarticletitle{Leveraging adversarial training in self-learning
  for cross-lingual text classification}. In
  \bibinfo{booktitle}{\emph{Proceedings of the 43rd International ACM SIGIR
  Conference on Research and Development in Information Retrieval}}.
  \bibinfo{pages}{1541--1544}.
\newblock


\bibitem[Guo et~al\mbox{.}(2021)]%
        {guo2021survey}
\bibfield{author}{\bibinfo{person}{Zhijiang Guo}, \bibinfo{person}{Michael
  Schlichtkrull}, {and} \bibinfo{person}{Andreas Vlachos}.}
  \bibinfo{year}{2021}\natexlab{}.
\newblock \showarticletitle{A survey on automated fact-checking}.
\newblock \bibinfo{journal}{\emph{arXiv preprint arXiv:2108.11896}}
  (\bibinfo{year}{2021}).
\newblock


\bibitem[Hasanain et~al\mbox{.}(2020)]%
        {clef-checkthat-ar:2020}
\bibfield{author}{\bibinfo{person}{Maram Hasanain}, \bibinfo{person}{Fatima
  Haouari}, \bibinfo{person}{Reem Suwaileh}, \bibinfo{person}{{Zien Sheikh}
  Ali}, \bibinfo{person}{Bayan Hamdan}, \bibinfo{person}{Tamer Elsayed},
  \bibinfo{person}{Alberto Barr\'{o}n-Cede{\~n}o}, \bibinfo{person}{Giovanni
  {Da San Martino}}, {and} \bibinfo{person}{Preslav Nakov}.}
  \bibinfo{year}{2020}\natexlab{}.
\newblock \showarticletitle{Overview of {CheckThat!} 2020 {A}rabic: Automatic
  Identification and Verification of Claims in Social Media}, See
  \citeN{clef2020-workingnotes}.
\newblock
\showISSN{1613-0073}


\bibitem[Hassan et~al\mbox{.}(2017)]%
        {Hassan2017}
\bibfield{author}{\bibinfo{person}{Naeemul Hassan}, \bibinfo{person}{Fatma
  Arslan}, \bibinfo{person}{Chengkai Li}, {and} \bibinfo{person}{Mark
  Tremayne}.} \bibinfo{year}{2017}\natexlab{}.
\newblock \showarticletitle{Toward Automated Fact-Checking: Detecting
  Check-Worthy Factual Claims by ClaimBuster}. In
  \bibinfo{booktitle}{\emph{Proceedings of the 23rd ACM SIGKDD International
  Conference on Knowledge Discovery and Data Mining}}
  \emph{(\bibinfo{series}{KDD '17})}. \bibinfo{publisher}{Association for
  Computing Machinery}, \bibinfo{pages}{1803–1812}.
\newblock
\showISBNx{9781450348874}


\bibitem[Hedderich et~al\mbox{.}(2020)]%
        {hedderich2020}
\bibfield{author}{\bibinfo{person}{Michael~A. Hedderich},
  \bibinfo{person}{David Adelani}, \bibinfo{person}{Dawei Zhu},
  \bibinfo{person}{Jesujoba Alabi}, \bibinfo{person}{Udia Markus}, {and}
  \bibinfo{person}{Dietrich Klakow}.} \bibinfo{year}{2020}\natexlab{}.
\newblock \showarticletitle{Transfer Learning and Distant Supervision for
  Multilingual Transformer Models: A Study on {A}frican Languages}. In
  \bibinfo{booktitle}{\emph{Proceedings of the 2020 Conference on Empirical
  Methods in Natural Language Processing (EMNLP)}}.
  \bibinfo{publisher}{Association for Computational Linguistics},
  \bibinfo{pages}{2580--2591}.
\newblock


\bibitem[Juan R. Martinez-Rico and Araujo(2021)]%
        {Martinez-Rico2021}
\bibfield{author}{\bibinfo{person}{Juan Martinez-Romo Juan R. Martinez-Rico}
  {and} \bibinfo{person}{Lourdes Araujo}.} \bibinfo{year}{2021}\natexlab{}.
\newblock \showarticletitle{{NLP\&IR@UNED} at {CheckThat!} 2021:
  Check-worthiness estimation and fake news detection using transformer
  models}, See \citeN{clef2021-workingnotes}.
\newblock


\bibitem[Kartal and Kutlu(2022)]%
        {Kartal2022}
\bibfield{author}{\bibinfo{person}{Yavuz~Selim Kartal} {and}
  \bibinfo{person}{Mucahid Kutlu}.} \bibinfo{year}{2022}\natexlab{}.
\newblock \showarticletitle{Re-Think Before You Share: A Comprehensive Study on
  Prioritizing Check-Worthy Claims}.
\newblock \bibinfo{journal}{\emph{IEEE Transactions on Computational Social
  Systems}} (\bibinfo{year}{2022}), \bibinfo{pages}{1--14}.
\newblock
\urldef\tempurl%
\url{https://doi.org/10.1109/TCSS.2021.3138642}
\showDOI{\tempurl}


\bibitem[Keung et~al\mbox{.}(2019)]%
        {keung2019}
\bibfield{author}{\bibinfo{person}{Phillip Keung}, \bibinfo{person}{Yichao Lu},
  {and} \bibinfo{person}{Vikas Bhardwaj}.} \bibinfo{year}{2019}\natexlab{}.
\newblock \showarticletitle{Adversarial Learning with Contextual Embeddings for
  Zero-resource Cross-lingual Classification and NER}. In
  \bibinfo{booktitle}{\emph{Proceedings of the 2019 Conference on Empirical
  Methods in Natural Language Processing and the 9th International Joint
  Conference on Natural Language Processing (EMNLP-IJCNLP)}}.
  \bibinfo{pages}{1355--1360}.
\newblock


\bibitem[Meng et~al\mbox{.}(2020)]%
        {meng2020gradient}
\bibfield{author}{\bibinfo{person}{Kevin Meng}, \bibinfo{person}{Damian
  Jimenez}, \bibinfo{person}{Fatma Arslan}, \bibinfo{person}{Jacob~Daniel
  Devasier}, \bibinfo{person}{Daniel Obembe}, {and} \bibinfo{person}{Chengkai
  Li}.} \bibinfo{year}{2020}\natexlab{}.
\newblock \showarticletitle{Gradient-based adversarial training on transformer
  networks for detecting check-worthy factual claims}.
\newblock \bibinfo{journal}{\emph{arXiv preprint arXiv:2002.07725}}
  (\bibinfo{year}{2020}).
\newblock


\bibitem[Nakov et~al\mbox{.}(2021a)]%
        {nakov2021automated}
\bibfield{author}{\bibinfo{person}{Preslav Nakov}, \bibinfo{person}{David
  Corney}, \bibinfo{person}{Maram Hasanain}, \bibinfo{person}{Firoj Alam},
  \bibinfo{person}{Tamer Elsayed}, \bibinfo{person}{Alberto Barrón-Cedeño},
  \bibinfo{person}{Paolo Papotti}, \bibinfo{person}{Shaden Shaar}, {and}
  \bibinfo{person}{Giovanni Da~San Martino}.} \bibinfo{year}{2021}\natexlab{a}.
\newblock \showarticletitle{Automated Fact-Checking for Assisting Human
  Fact-Checkers} \emph{(\bibinfo{series}{IJCAI-2021})}.
\newblock


\bibitem[Nakov et~al\mbox{.}(2021b)]%
        {clef-checkthat:2021:LNCS}
\bibfield{author}{\bibinfo{person}{Preslav Nakov},
  \bibinfo{person}{Da~San~Martino Giovanni}, \bibinfo{person}{Tamer Elsayed},
  \bibinfo{person}{Alberto Barr{\'{o}}n{-}Cede{\~{n}}o},
  \bibinfo{person}{Rub\'{e}n M\'{i}guez}, \bibinfo{person}{Shaden Shaar},
  \bibinfo{person}{Firoj Alam}, \bibinfo{person}{Fatima Haouari},
  \bibinfo{person}{Maram Hasanain}, \bibinfo{person}{Watheq Mansour},
  \bibinfo{person}{Bayan Hamdan}, \bibinfo{person}{Zien~Sheikh Ali},
  \bibinfo{person}{Nikolay Babulkov}, \bibinfo{person}{Alex Nikolov},
  \bibinfo{person}{Gautam~Kishore Shahi}, \bibinfo{person}{Julia~Maria Struß},
  \bibinfo{person}{Thomas Mandl}, \bibinfo{person}{Mucahid Kutlu}, {and}
  \bibinfo{person}{Yavuz~Selim Kartal}.} \bibinfo{year}{2021}\natexlab{b}.
\newblock \showarticletitle{Overview of the {CLEF}-2021 {CheckThat}! Lab on
  Detecting Check-Worthy Claims, Previously Fact-Checked Claims, and Fake News}
  \emph{(\bibinfo{series}{{CLEF} 2021})},
  \bibfield{editor}{\bibinfo{person}{{K. Selcuk} Candan},
  \bibinfo{person}{Bogdan Ionescu}, \bibinfo{person}{Lorraine Goeuriot},
  \bibinfo{person}{Birger Larsen}, \bibinfo{person}{Henning Müller},
  \bibinfo{person}{Alexis Joly}, \bibinfo{person}{Maria Maistro},
  \bibinfo{person}{Florina Piroi}, \bibinfo{person}{Guglielmo Faggioli}, {and}
  \bibinfo{person}{Nicola Ferro}} (Eds.). \bibinfo{publisher}{Springer}.
\newblock


\bibitem[Panda and Levitan(2021)]%
        {panda2021detecting}
\bibfield{author}{\bibinfo{person}{Subhadarshi Panda} {and}
  \bibinfo{person}{Sarah~Ita Levitan}.} \bibinfo{year}{2021}\natexlab{}.
\newblock \showarticletitle{Detecting multilingual COVID-19 misinformation on
  social media via contextualized embeddings}. In
  \bibinfo{booktitle}{\emph{Proceedings of the Fourth Workshop on NLP for
  Internet Freedom: Censorship, Disinformation, and Propaganda}}.
  \bibinfo{pages}{125--129}.
\newblock


\bibitem[Pires et~al\mbox{.}(2019)]%
        {pires2019}
\bibfield{author}{\bibinfo{person}{Telmo Pires}, \bibinfo{person}{Eva
  Schlinger}, {and} \bibinfo{person}{Dan Garrette}.}
  \bibinfo{year}{2019}\natexlab{}.
\newblock \showarticletitle{How Multilingual is Multilingual BERT?}. In
  \bibinfo{booktitle}{\emph{Proceedings of the 57th Annual Meeting of the
  Association for Computational Linguistics}}. \bibinfo{pages}{4996--5001}.
\newblock


\bibitem[Reimers and Gurevych(2020)]%
        {reimers2020making}
\bibfield{author}{\bibinfo{person}{Nils Reimers} {and} \bibinfo{person}{Iryna
  Gurevych}.} \bibinfo{year}{2020}\natexlab{}.
\newblock \showarticletitle{Making Monolingual Sentence Embeddings Multilingual
  Using Knowledge Distillation}. In \bibinfo{booktitle}{\emph{Proceedings of
  the 2020 Conference on Empirical Methods in Natural Language Processing
  (EMNLP)}}. \bibinfo{pages}{4512--4525}.
\newblock


\bibitem[Samarinas et~al\mbox{.}(2021)]%
        {samarinas-etal-2021-improving}
\bibfield{author}{\bibinfo{person}{Chris Samarinas}, \bibinfo{person}{Wynne
  Hsu}, {and} \bibinfo{person}{Mong~Li Lee}.} \bibinfo{year}{2021}\natexlab{}.
\newblock \showarticletitle{Improving Evidence Retrieval for Automated
  Explainable Fact-Checking}. In \bibinfo{booktitle}{\emph{Proceedings of the
  2021 Conference of the North American Chapter of the Association for
  Computational Linguistics: Human Language Technologies: Demonstrations}}.
  \bibinfo{publisher}{Association for Computational Linguistics},
  \bibinfo{address}{Online}, \bibinfo{pages}{84--91}.
\newblock
\urldef\tempurl%
\url{https://doi.org/10.18653/v1/2021.naacl-demos.10}
\showDOI{\tempurl}


\bibitem[Schwenk and Li(2018)]%
        {SCHWENK18.658}
\bibfield{author}{\bibinfo{person}{Holger Schwenk} {and} \bibinfo{person}{Xian
  Li}.} \bibinfo{year}{2018}\natexlab{}.
\newblock \showarticletitle{A Corpus for Multilingual Document Classification
  in Eight Languages}. In \bibinfo{booktitle}{\emph{Proceedings of the Eleventh
  International Conference on Language Resources and Evaluation (LREC 2018)}}
  (Miyazaki, Japan, 7-12), \bibfield{editor}{\bibinfo{person}{Nicoletta
  Calzolari~(Conference chair)}, \bibinfo{person}{Khalid Choukri},
  \bibinfo{person}{Christopher Cieri}, \bibinfo{person}{Thierry Declerck},
  \bibinfo{person}{Sara Goggi}, \bibinfo{person}{Koiti Hasida},
  \bibinfo{person}{Hitoshi Isahara}, \bibinfo{person}{Bente Maegaard},
  \bibinfo{person}{Joseph Mariani}, \bibinfo{person}{Hélène Mazo},
  \bibinfo{person}{Asuncion Moreno}, \bibinfo{person}{Jan Odijk},
  \bibinfo{person}{Stelios Piperidis}, {and} \bibinfo{person}{Takenobu
  Tokunaga}} (Eds.). \bibinfo{publisher}{European Language Resources
  Association (ELRA)}, \bibinfo{address}{Paris, France}.
\newblock
\showISBNx{979-10-95546-00-9}


\bibitem[Schweter(2020)]%
        {stefan_schweter_2020_3770924}
\bibfield{author}{\bibinfo{person}{Stefan Schweter}.}
  \bibinfo{year}{2020}\natexlab{}.
\newblock \bibinfo{booktitle}{\emph{BERTurk - BERT models for Turkish}}.
\newblock
\urldef\tempurl%
\url{https://doi.org/10.5281/zenodo.3770924}
\showDOI{\tempurl}


\bibitem[Sear et~al\mbox{.}(2020)]%
        {sear2020quantifying}
\bibfield{author}{\bibinfo{person}{Richard~F Sear},
  \bibinfo{person}{Nicol{\'a}s Vel{\'a}squez}, \bibinfo{person}{Rhys Leahy},
  \bibinfo{person}{Nicholas~Johnson Restrepo}, \bibinfo{person}{Sara El~Oud},
  \bibinfo{person}{Nicholas Gabriel}, \bibinfo{person}{Yonatan Lupu}, {and}
  \bibinfo{person}{Neil~F Johnson}.} \bibinfo{year}{2020}\natexlab{}.
\newblock \showarticletitle{Quantifying COVID-19 content in the online health
  opinion war using machine learning}.
\newblock \bibinfo{journal}{\emph{Ieee Access}}  \bibinfo{volume}{8}
  (\bibinfo{year}{2020}), \bibinfo{pages}{91886--91893}.
\newblock


\bibitem[Sep\'ulveda-Torres and Saquete(2021)]%
        {Sepulveda2021}
\bibfield{author}{\bibinfo{person}{Robiert Sep\'ulveda-Torres} {and}
  \bibinfo{person}{Estela Saquete}.} \bibinfo{year}{2021}\natexlab{}.
\newblock \showarticletitle{{GPLSI} team at {CLEF CheckThat! 2021}: Fine-tuning
  {BETO} and {RoBERTa}}, See \citeN{clef2021-workingnotes}.
\newblock


\bibitem[Shaar et~al\mbox{.}(2021a)]%
        {shaar-etal-2021-findings}
\bibfield{author}{\bibinfo{person}{Shaden Shaar}, \bibinfo{person}{Firoj Alam},
  \bibinfo{person}{Giovanni Da~San~Martino}, \bibinfo{person}{Alex Nikolov},
  \bibinfo{person}{Wajdi Zaghouani}, \bibinfo{person}{Preslav Nakov}, {and}
  \bibinfo{person}{Anna Feldman}.} \bibinfo{year}{2021}\natexlab{a}.
\newblock \showarticletitle{Findings of the {NLP}4{IF}-2021 Shared Tasks on
  Fighting the {COVID}-19 Infodemic and Censorship Detection}. In
  \bibinfo{booktitle}{\emph{Proceedings of the Fourth Workshop on NLP for
  Internet Freedom: Censorship, Disinformation, and Propaganda}}.
  \bibinfo{publisher}{Association for Computational Linguistics},
  \bibinfo{address}{Online}, \bibinfo{pages}{82--92}.
\newblock
\urldef\tempurl%
\url{https://doi.org/10.18653/v1/2021.nlp4if-1.12}
\showDOI{\tempurl}


\bibitem[Shaar et~al\mbox{.}(2021b)]%
        {clef-checkthat:2021:task1}
\bibfield{author}{\bibinfo{person}{Shaden Shaar}, \bibinfo{person}{Maram
  Hasanain}, \bibinfo{person}{Bayan Hamdan}, \bibinfo{person}{Zien~Sheikh Ali},
  \bibinfo{person}{Fatima Haouari}, \bibinfo{person}{Alex Nikolov},
  \bibinfo{person}{Mucahid Kutlu}, \bibinfo{person}{Yavuz~Selim Kartal},
  \bibinfo{person}{Firoj Alam}, \bibinfo{person}{Giovanni Da~San~Martino},
  \bibinfo{person}{Alberto Barr{\'{o}}n{-}Cede{\~{n}}o},
  \bibinfo{person}{Rub\'{e}n M\'{i}guez}, \bibinfo{person}{Javier Beltr\'an},
  \bibinfo{person}{Tamer Elsayed}, {and} \bibinfo{person}{Preslav Nakov}.}
  \bibinfo{year}{2021}\natexlab{b}.
\newblock \showarticletitle{Overview of the {CLEF}-2021 {CheckThat}! Lab Task 1
  on Check-Worthiness Estimation in Tweets and Political Debates}, See
  \citeN{clef2021-workingnotes}.
\newblock


\bibitem[Shaar et~al\mbox{.}(2020)]%
        {clef-checkthat-en:2020}
\bibfield{author}{\bibinfo{person}{Shaden Shaar}, \bibinfo{person}{Alex
  Nikolov}, \bibinfo{person}{Nikolay Babulkov}, \bibinfo{person}{Firoj Alam},
  \bibinfo{person}{Alberto Barr\'{o}n-Cede{\~n}o}, \bibinfo{person}{Tamer
  Elsayed}, \bibinfo{person}{Maram Hasanain}, \bibinfo{person}{Reem Suwaileh},
  \bibinfo{person}{Fatima Haouari}, \bibinfo{person}{Giovanni {Da San
  Martino}}, {and} \bibinfo{person}{Preslav Nakov}.}
  \bibinfo{year}{2020}\natexlab{}.
\newblock \showarticletitle{Overview of {CheckThat!} 2020 {E}nglish: Automatic
  Identification and Verification of Claims in Social Media}, See
  \citeN{clef2020-workingnotes}.
\newblock
\showISSN{1613-0073}


\bibitem[Uyangodage et~al\mbox{.}(2021)]%
        {uyangodage2021}
\bibfield{author}{\bibinfo{person}{Lasitha Uyangodage},
  \bibinfo{person}{Tharindu Ranasinghe}, {and} \bibinfo{person}{Hansi
  Hettiarachchi}.} \bibinfo{year}{2021}\natexlab{}.
\newblock \showarticletitle{Can Multilingual Transformers Fight the COVID-19
  Infodemic}. In \bibinfo{booktitle}{\emph{Proceedings of RANLP}}.
\newblock


\bibitem[Williams et~al\mbox{.}(2021a)]%
        {clef-checkthat:2021:task1:accenture}
\bibfield{author}{\bibinfo{person}{Evan Williams}, \bibinfo{person}{Paul
  Rodrigues}, {and} \bibinfo{person}{Sieu Tran}.}
  \bibinfo{year}{2021}\natexlab{a}.
\newblock \showarticletitle{Accenture at {CheckThat!} 2021: Interesting claim
  identification and ranking with contextually sensitive lexical training data
  augmentation}, See \citeN{clef2021-workingnotes}.
\newblock


\bibitem[Williams et~al\mbox{.}(2021b)]%
        {accenture2021}
\bibfield{author}{\bibinfo{person}{Evan Williams}, \bibinfo{person}{Paul
  Rodrigues}, {and} \bibinfo{person}{Sieu Tran}.}
  \bibinfo{year}{2021}\natexlab{b}.
\newblock \showarticletitle{Accenture at {CheckThat!} 2021: Interesting claim
  identification and ranking with contextually sensitive lexical training data
  augmentation}, See \citeN{clef2021-workingnotes}.
\newblock


\bibitem[Wright and Augenstein(2020)]%
        {wright2020claim}
\bibfield{author}{\bibinfo{person}{Dustin Wright} {and}
  \bibinfo{person}{Isabelle Augenstein}.} \bibinfo{year}{2020}\natexlab{}.
\newblock \showarticletitle{Claim Check-Worthiness Detection as Positive
  Unlabelled Learning}. In \bibinfo{booktitle}{\emph{Proceedings of the 2020
  Conference on Empirical Methods in Natural Language Processing: Findings}}.
  \bibinfo{pages}{476--488}.
\newblock


\bibitem[Wu and Dredze(2019)]%
        {wu2019beto}
\bibfield{author}{\bibinfo{person}{Shijie Wu} {and} \bibinfo{person}{Mark
  Dredze}.} \bibinfo{year}{2019}\natexlab{}.
\newblock \showarticletitle{Beto, Bentz, Becas: The Surprising Cross-Lingual
  Effectiveness of {BERT}}. In \bibinfo{booktitle}{\emph{Proceedings of the
  2019 Conference on Empirical Methods in Natural Language Processing and the
  9th International Joint Conference on Natural Language Processing
  (EMNLP-IJCNLP)}}. \bibinfo{publisher}{Association for Computational
  Linguistics}, \bibinfo{address}{Hong Kong, China}, \bibinfo{pages}{833--844}.
\newblock
\urldef\tempurl%
\url{https://doi.org/10.18653/v1/D19-1077}
\showDOI{\tempurl}


\bibitem[Wu and Dredze(2020)]%
        {wu2020all}
\bibfield{author}{\bibinfo{person}{Shijie Wu} {and} \bibinfo{person}{Mark
  Dredze}.} \bibinfo{year}{2020}\natexlab{}.
\newblock \showarticletitle{Are All Languages Created Equal in Multilingual
  BERT?}. In \bibinfo{booktitle}{\emph{Proceedings of the 5th Workshop on
  Representation Learning for NLP}}. \bibinfo{pages}{120--130}.
\newblock


\bibitem[Zeng et~al\mbox{.}(2021)]%
        {zeng:survey:2021}
\bibfield{author}{\bibinfo{person}{Xia Zeng}, \bibinfo{person}{Amani~S.
  Abumansour}, {and} \bibinfo{person}{Arkaitz Zubiaga}.}
  \bibinfo{year}{2021}\natexlab{}.
\newblock \showarticletitle{Automated fact-checking: A survey}.
\newblock \bibinfo{journal}{\emph{Language and Linguistics Compass}}
  \bibinfo{volume}{15}, \bibinfo{number}{10} (\bibinfo{year}{2021}),
  \bibinfo{pages}{e12438}.
\newblock
\urldef\tempurl%
\url{https://doi.org/10.1111/lnc3.12438}
\showDOI{\tempurl}
\showeprint{https://onlinelibrary.wiley.com/doi/pdf/10.1111/lnc3.12438}


\bibitem[Zengin et~al\mbox{.}(2021)]%
        {tobbetu2021}
\bibfield{author}{\bibinfo{person}{Muhammed~Said Zengin},
  \bibinfo{person}{Yavuz~Selim Kartal}, {and} \bibinfo{person}{Mucahid Kutlu}.}
  \bibinfo{year}{2021}\natexlab{}.
\newblock \showarticletitle{{TOBB ETU} at {CheckThat! 2021}: Data Engineering
  for Detecting Check-Worthy Claims}, See \citeN{clef2021-workingnotes}.
\newblock


\bibitem[Zhao et~al\mbox{.}(2021)]%
        {zhao2021closer}
\bibfield{author}{\bibinfo{person}{Mengjie Zhao}, \bibinfo{person}{Yi Zhu},
  \bibinfo{person}{Ehsan Shareghi}, \bibinfo{person}{Ivan Vuli{\'c}},
  \bibinfo{person}{Roi Reichart}, \bibinfo{person}{Anna Korhonen}, {and}
  \bibinfo{person}{Hinrich Sch{\"u}tze}.} \bibinfo{year}{2021}\natexlab{}.
\newblock \showarticletitle{A closer look at few-shot crosslingual transfer:
  The choice of shots matters}. In \bibinfo{booktitle}{\emph{Proceedings of the
  59th Annual Meeting of the Association for Computational Linguistics and the
  11th International Joint Conference on Natural Language Processing (Volume 1:
  Long Papers)}}. \bibinfo{pages}{5751--5767}.
\newblock


\end{thebibliography}

\end{document}